\documentclass{article}
\PassOptionsToPackage{table}{xcolor}
\usepackage{iclr2027_conference,times}
\usepackage{multirow}
\usepackage{booktabs}
\usepackage{graphicx}
\usepackage{wrapfig}
\usepackage{amsmath}
\usepackage{algorithm}
\usepackage{algpseudocode}
\usepackage{caption}

\usepackage{amsmath,amsfonts,bm}

\def\eqref#1{equation~\ref{#1}}
\def\1{\bm{1}}

\DeclareMathAlphabet{\mathsfit}{\encodingdefault}{\sfdefault}{m}{sl}
\SetMathAlphabet{\mathsfit}{bold}{\encodingdefault}{\sfdefault}{bx}{n}

\title{\textsc{Collator}: Compositional Multi-Agent Orchestration with Counterfactual Reinforcement Learning}

\author{
Xudong Chen \quad
Yixin Liu\textsuperscript{1} \quad
Hua Wei\textsuperscript{2} \quad
Kaize Ding\textsuperscript{3} \\[0.5em]
\normalfont\textsuperscript{1} Griffith University \\
\normalfont\textsuperscript{2} Arizona State University \\
\normalfont\textsuperscript{3} Northwestern University
}

\usepackage[utf8]{inputenc}
\usepackage[T1]{fontenc}
\usepackage{hyperref}
\usepackage{url}
\usepackage{amsfonts}
\usepackage{nicefrac}
\usepackage{microtype}
\usepackage{xcolor}
\usepackage{enumitem}
\usepackage{xspace}
\definecolor{BestPurple}{RGB}{232,222,248}
\definecolor{SecondOrange}{RGB}{255,235,214}

\newcommand{\methodname}{\textsc{Collator}\xspace}

\usepackage[most]{tcolorbox}
\usepackage{listings}

\definecolor{PromptGreen}{RGB}{0,155,85}
\definecolor{PromptShadow}{RGB}{190,190,190}
\definecolor{PromptBg}{RGB}{250,250,250}

\iclrfinalcopy

\begin{document}

\maketitle
\lhead{Preprint}

\newcommand{\upabs}[1]{{\tiny\,\ensuremath{\uparrow #1}}}
\newcommand{\downabs}[1]{{\tiny\,\ensuremath{\downarrow #1}}}

\newcommand{\mainresultstable}{
\begin{table*}[t]
\vspace{-8pt}
\centering
\caption{
Performance comparison (\%) across six benchmarks.
We report answer-option accuracy for MMLU and AQuA, exact-match accuracy for GSM8K, MultiArith, and SVAMP, and pass@1 for HumanEval.
Avg. is the unweighted arithmetic mean of the six reported percentages.
}\label{tab:main_results}
\resizebox{\textwidth}{!}{
\begin{tabular}{lccccccc}
\toprule
\textbf{Method}
& \textbf{MMLU$^{\dagger}$}
& \textbf{GSM8K}
& \textbf{AQuA}
& \textbf{MultiArith}
& \textbf{SVAMP}
& \textbf{HumanEval$^{\ddagger}$}
& \textbf{Avg.} \\
\midrule
\rowcolor{gray!8}
\multicolumn{8}{l}{\textit{Single-Agent Reasoning}} \\[-1pt]
Vanilla
& 74.46 & 85.59 & 46.46 & 76.00 & 80.80 & 80.64 & 73.99 \\
CoT
& 66.49\downabs{7.97} & 91.36\upabs{5.77} & 81.89\upabs{35.43} & 98.67\upabs{22.67} & 93.60\upabs{12.80} & 81.45\upabs{0.81} & 85.58\upabs{11.59} \\
SC-CoT
& 67.54\downabs{6.92} & 93.25\upabs{7.66} & 82.68\upabs{36.22} & 98.83\upabs{22.83} & 94.40\upabs{13.60} & \cellcolor{SecondOrange}\underline{83.88}\upabs{3.24} & 86.76\upabs{12.77} \\
\midrule
\rowcolor{gray!8}
\multicolumn{8}{l}{\textit{Fixed Multi-Agent Orchestration}} \\[-1pt]
Chain-MAS
& 78.51\upabs{4.05} & 92.19\upabs{6.60} & 77.17\upabs{30.71} & 98.83\upabs{22.83} & 94.40\upabs{13.60} & 80.65\upabs{0.01} & 86.96\upabs{12.97} \\
Tree-MAS
& 79.56\upabs{5.10} & 92.49\upabs{6.90} & 82.68\upabs{36.22} & 98.67\upabs{22.67} & 93.90\upabs{13.10} & \cellcolor{SecondOrange}\underline{83.88}\upabs{3.24} & \cellcolor{SecondOrange}\underline{88.53}\upabs{14.54} \\
Complete-MAS
& 79.16\upabs{4.70} & 92.65\upabs{7.06} & 83.46\upabs{37.00} & 98.83\upabs{22.83} & 94.00\upabs{13.20} & 82.25\upabs{1.61} & 88.39\upabs{14.40} \\
Random-MAS
& 79.69\upabs{5.23} & 92.19\upabs{6.60} & 81.10\upabs{34.64} & 98.67\upabs{22.67} & 94.00\upabs{13.20} & 83.06\upabs{2.42} & 88.12\upabs{14.13} \\
\midrule
\rowcolor{gray!8}
\multicolumn{8}{l}{\textit{Adaptive Workflow Orchestration}} \\[-1pt]
AFlow
& 70.35\downabs{4.11} & 91.05\upabs{5.46} & \cellcolor{SecondOrange}\underline{83.86}\upabs{37.40} & 96.33\upabs{20.33} & 93.80\upabs{13.00} & 76.61\downabs{4.03} & 85.33\upabs{11.34} \\
\midrule
\rowcolor{gray!8}
\multicolumn{8}{l}{\textit{Topology Design}} \\[-1pt]
AgentDropout
& 80.73\upabs{6.27} & 92.87\upabs{7.28} & 83.07\upabs{36.61} & 97.50\upabs{21.50} & 93.10\upabs{12.30} & 77.42\downabs{3.22} & 87.45\upabs{13.46} \\
AgentPrune
& 80.54\upabs{6.08} & 91.36\upabs{5.77} & \cellcolor{SecondOrange}\underline{83.86}\upabs{37.40} & 98.00\upabs{22.00} & \cellcolor{SecondOrange}\underline{95.20}\upabs{14.40} & 79.84\downabs{0.80} & 88.13\upabs{14.14} \\
G-Designer
& 78.25\upabs{3.79} & 92.17\upabs{6.58} & 82.28\upabs{35.82} & 98.17\upabs{22.17} & 93.10\upabs{12.30} & 71.77\downabs{8.87} & 85.96\upabs{11.97} \\
ARG-Designer
& \cellcolor{BestPurple}\textbf{85.30}\upabs{10.84} & \cellcolor{SecondOrange}\underline{93.35}\upabs{7.76} & 69.29\upabs{22.83} & \cellcolor{SecondOrange}\underline{99.00}\upabs{23.00} & 94.90\upabs{14.10} & 76.61\downabs{4.03} & 86.41\upabs{12.42} \\
OFA-MAS
& 78.38\upabs{3.92} & 84.84\downabs{0.75} & 82.67\upabs{36.21} & 98.67\upabs{22.67} & 94.20\upabs{13.40} & 79.84\downabs{0.80} & 86.43\upabs{12.44} \\
\midrule
\rowcolor{gray!8}
\multicolumn{8}{l}{\textit{Compositional Orchestration Design}} \\[-1pt]
\textbf{Ours}
& \cellcolor{SecondOrange}\underline{84.32}\upabs{9.86} & \cellcolor{BestPurple}\textbf{94.32}\upabs{8.73} & \cellcolor{BestPurple}\textbf{86.22}\upabs{39.76} & \cellcolor{BestPurple}\textbf{99.17}\upabs{23.17} & \cellcolor{BestPurple}\textbf{95.60}\upabs{14.80} & \cellcolor{BestPurple}\textbf{84.68}\upabs{4.04} & \cellcolor{BestPurple}\textbf{90.72}\upabs{16.73} \\
\bottomrule
\end{tabular}
}
\begin{minipage}{\textwidth}
\footnotesize
$^{\dagger}$MMLU is evaluated on the validation set with 1,531 examples rather than the full test set.
$^{\ddagger}$HumanEval is evaluated on the held-out 124-example split for all methods; the remaining 40 examples are used only for training/adaptation by methods that require it. Full evaluation protocol is listed in Appendix~\ref{app:evaluation_protocol}.
\end{minipage}
\vspace{-12pt}
\end{table*}
}

\newcommand{\ablationtable}{
\begin{wraptable}{r}{0.55\textwidth}
\centering
\scriptsize
\setlength{\tabcolsep}{2.5pt}
\vspace{-8pt}
\caption{Ablation studies on representative benchmarks.}
\label{tab:ablation}
\resizebox{\linewidth}{!}{
\begin{tabular}{lcccccc}
\toprule
\textbf{Method}
& \multicolumn{2}{c}{\textbf{MMLU}}
& \multicolumn{2}{c}{\textbf{GSM8K}}
& \multicolumn{2}{c}{\textbf{AQuA}} \\
\cmidrule(lr){2-3}
\cmidrule(lr){4-5}
\cmidrule(lr){6-7}
& \textbf{Acc.}
& \textbf{Tok.$\downarrow$}
& \textbf{Acc.}
& \textbf{Tok.$\downarrow$}
& \textbf{Acc.}
& \textbf{Tok.$\downarrow$} \\
\midrule
SFT only
& 72.89 & 1950.51
& 81.65 & 1564.01
& 79.53 & 1421.90 \\
SFT + GRPO
& 83.08 & 2543.21
& 93.40 & 3742.59
& 85.43 & 3012.54 \\
\textbf{Ours}
& \textbf{84.32} & \textbf{2148.00}
& \textbf{94.32} & \textbf{3362.00}
& \textbf{86.22} & \textbf{2056.10} \\
\midrule
w/o Role Design
& 82.43 & 2496.77
& 92.72 & 3734.76
& 83.07 & 2829.91 \\
w/o Cap. Assignment
& 82.76 & 3401.06
& 93.18 & 4310.98
& 84.65 & 4154.38 \\
w/o Dep. Construction
& 80.86 & 2347.71
& 93.33 & 3991.04
& 83.86 & 3296.94 \\
\midrule
w/o Dep. CF
& 83.54 & 2516.13
& 93.78 & 3797.33
& 85.43 & 3303.37 \\
w/o Role CF
& 83.67 & 2484.87
& 93.93 & 3553.81
& 85.83 & 2680.48 \\
w/o Cap. CF
& 84.00 & 2663.06
& 94.09 & 4362.87
& 85.83 & 3589.65 \\
\midrule
w/o SFT
& 48.53 & 2777.60
& 75.82 & 3964.21
& 64.96 & 3729.20 \\
\bottomrule
\vspace{-12pt}
\end{tabular}
}
\end{wraptable}
}
\begin{abstract}
Large language models (LLMs) provide a flexible foundation for multi-agent systems, but their effectiveness and computational cost depend critically on orchestration design. Across different tasks, role design, capacity assignment, and dependency construction jointly affect both solution quality and execution efficiency. Existing approaches automate parts of this design process, yet they often optimize these decisions partially or sequentially, and rely on execution-level feedback that provides limited credit assignment for local orchestration decisions. We propose \methodname{} (Compositional Multi-Agent Orchestration with Counterfactual Reinforcement Learning), an LLM-based orchestrator that learns to design efficient multi-agent orchestration. Given a task, \methodname{} designs a unified orchestration specification that composes customized agent duties, capacity levels, and dependency relations. To train the orchestrator, we augment the orchestration-level Group Relative Policy Optimization (GRPO) objective with a localized counterfactual credit signal that edits role, capacity, or dependency fields and applies the resulting reward contrast only to the edited spans. Experiments on six reasoning and coding benchmarks, including MMLU, GSM8K, AQuA, MultiArith, SVAMP, and HumanEval, show that \methodname{} achieves the best average performance among evaluated multi-agent orchestration methods while improving the accuracy--token trade-off. Our code is available at \url{https://anonymous.4open.science/r/LEMON-B23C}.
\end{abstract}

\section{Introduction}
Recent advances in large language models (LLMs) have driven growing interest in LLM-based multi-agent systems (MAS)~\cite{wu2024autogen, li2023camel, hong2023metagpt}. By coordinating multiple agents, these systems can leverage tool use, decomposition, verification, and collaborative reasoning to tackle tasks that are difficult for a single agent.~\cite{yao2022react, shinn2023reflexion}
However, their effectiveness depends not only on the capability of individual agents but also on how the system is orchestrated~\cite{guo2024large}. In particular, role specialization, capacity assignment, and information flow jointly determine how well the agents collaborate.~\cite{zhuge2024gptswarm, zhang2024cut} A fixed orchestration design is hard to reuse across tasks: the same design can waste computation on simple queries, be too weak for harder ones, and become mismatched when the required information flow changes~\cite{du2024improving, yao2023tree}. Together, these constraints make hand-designed, task-specific orchestration difficult to scale, and point to the need for automatic methods that adapt orchestration to each task.

Recent work has begun to move MAS orchestration beyond fixed manual designs by adapting different parts of the orchestration. Topology-learning methods~\cite{zhuge2024gptswarm, zhang2024g, zhang2024cut} optimize information-flow structures, but usually over a specified set of agents or role profiles. Workflow-orchestration methods learn executable procedures for decomposing tasks, invoking workers, and routing intermediate context~\cite{zhang2024aflow, nielsen2025learning, fan2024workflowllm}, while largely inheriting worker definitions and model capacity assignment from the available workflow space. Routing and team-selection methods further adapt role assignment, architecture search, or model routing~\cite{zhang2025multi, cai2025agentbalance}, but often connect these decisions through staged or cascaded procedures rather than one unified orchestration specification. 
Despite these methods making MAS orchestration more learnable and adaptive, this field remains in its infancy.

\begin{wrapfigure}{r}{0.6\linewidth}
\vspace{-8pt}
    \centering
    \includegraphics[width=1\linewidth]{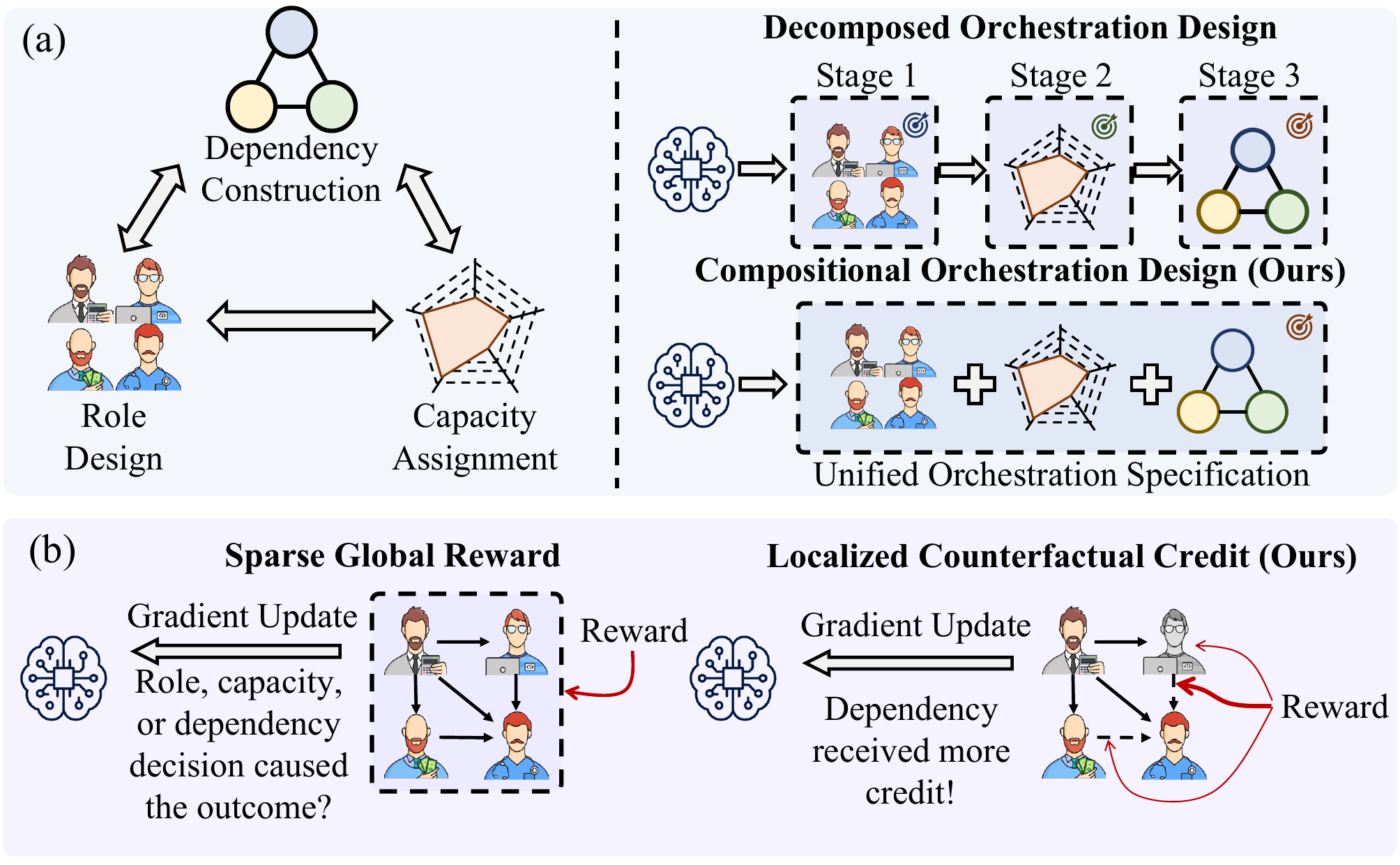}
    \caption{Compared with decomposed orchestration design and sparse global reward, \methodname{} composes role, capacity, and dependency fields within a unified orchestration specification and localizes counterfactual credit to the edited field.}
    \label{fig:intro}
\vspace{-12pt}
\end{wrapfigure}

The main limitation is that existing methods often optimize only part of the orchestration design space at the level of individual design components, rather than designing a unified orchestration specification that defines the multi-agent system compositionally~\cite{zhuge2024gptswarm,zhang2024g,zhang2024cut,zhang2024aflow,zhang2025multi,cai2025agentbalance}. As illustrated in Figure~\ref{fig:intro}(a), role design, capacity assignment, and dependency construction are coupled orchestration components, yet prior methods often determine them through decomposed stage-wise decisions~\cite{zhang2024aflow,nielsen2025learning,fan2024workflowllm,zhang2025multi}. Such decomposition separates task-specific agent duties, capacity levels, and dependency relations from a single specification-level design process, even though these choices interact during execution and jointly determine the resulting solution quality and computational cost. This motivates the problem of compositional multi-agent orchestration, where the goal is to design a task-adaptive unified specification that treats these orchestration fields as mutually dependent.

Given this formulation, learning compositional orchestration specifications introduces two challenges: \textbf{(i) Resource efficiency.} The orchestrator must decide how much computation each task deserves, since a more elaborate specification is not always a better one. Introducing more agents or dependencies can improve robustness, but it also raises two coupled costs: higher worker-token usage from running more agents, and higher structural cost, i.e., the overhead of a more complex orchestration structure~\cite{qian2025scaling,zhang2024cut,wang2025agentdropout}. In particular, denser dependencies increase inter-agent communication, so more intermediate outputs are passed as context and even more worker tokens are consumed~\cite{zhang2024cut,wang2025agentdropout}. An effective specification should therefore add computation only when the task actually benefits from it; \textbf{(ii) Sparse credit assignment.} Most training feedback for LLM-based orchestrators is derived from end-to-end execution outcomes~\cite{zhang2024aflow,nielsen2025learning}, such as task success or the overall utility of the executed specification. Such orchestration-level feedback can evaluate whether a specification succeeds as a whole, but it does not reveal which local role, capacity, or dependency decision is responsible for the outcome. As shown in Figure~\ref{fig:intro}(b), a sparse global reward assigns only one coarse training signal to the whole specification, leaving it unclear whether the outcome should be attributed to a role specialization, a capacity choice, or a dependency relation.

%\kz{you never show why this name}
To address these issues, we propose \methodname{}(Compositional Multi-Agent Orchestration with Counterfactual Reinforcement Learning), an LLM-based framework for learning efficient multi-agent orchestration. \methodname{} formulates MAS orchestration as \textbf{compositional orchestration design}: given an input task, the orchestrator produces a structured specification that composes task-specific agent duties, capacity levels, and information dependencies into a unified orchestration, thereby shifting learning from isolated component optimization, such as topology search, workflow construction, or routing, to specification-level orchestration design. To train the orchestrator, \methodname{} introduces \textbf{localized counterfactual reinforcement learning}. For each generated specification, \methodname{} edits one role, capacity, or dependency field to construct a counterfactual specification, executes it, and compares its execution reward with that of the original specification. Rather than exhaustively evaluating all possible local edits, \methodname{} uses sampled local counterfactuals to provide localized credit while keeping the additional training-time overhead manageable. The resulting reward contrast is mapped back only to the edited span, changing the granularity of credit assignment from the complete specification to the local orchestration field. Our contributions are summarized as follows:

\begin{itemize}[noitemsep,leftmargin=*,topsep=1.5pt]
    \item \textbf{Compositional orchestration design.}
    We formulate LLM-based MAS orchestration as designing a unified orchestration specification that composes task-specific role specifications, capacity levels, and dependency references, and can be directly validated, compiled, and executed.
    
    \item \textbf{Localized counterfactual reinforcement learning.} We address the sparse credit-assignment problem in orchestration learning by refining orchestration-level GRPO with a localized counterfactual credit signal. This signal converts execution-level reward differences into span-level supervision for the edited orchestration decision. 

    \item \textbf{Empirical validation.}
    Experiments across reasoning and coding benchmarks show that \methodname{} improves over single-agent, fixed-structure MAS, topology-design, and adaptive workflow baselines, with ablations validating both compositional orchestration design and localized counterfactual supervision.
\end{itemize}
\section{Related Work}
 \noindent \textbf{Multi-Agent Collaboration and Orchestration Design.}
Recent work has moved LLM-based multi-agent systems (MAS) beyond fixed manual designs by automating different components of orchestration. Topology-oriented methods learn or search communication structures, e.g., sparse communication, graph pruning, or task-adaptive graph generation, but usually operate over predefined agents or role profiles \cite{zhuge2024gptswarm, zhang2024cut, zhang2024g, li2026assemble, li2026ofa}. Workflow-oriented methods search or generate agentic procedures for task decomposition, worker invocation, and context routing, but typically rely on predefined worker interfaces, operators, or workflow primitives \cite{hu2024automated, zhang2024aflow, nielsen2025learning, su2026difficulty}. Routing and team-construction methods adapt collaboration modes, agent sets, or heterogeneous LLM backbones, but often determine these choices through staged or cascaded controllers \cite{zhang2025multi, yue2025masrouter, cai2025agentbalance, yao2026hieramas}. Overall, these methods improve the adaptivity of multi-agent orchestration, but they typically optimize topology, workflow, routing, or model selection as separate components rather than composing agent duties, capacity levels, and dependency relations into a unified orchestration specification. In contrast, we formulate orchestration as compositional orchestration design, where agent duties, capacity levels, and dependency relations are specified together in the same orchestration specification that is directly validated, compiled, and executed.

\noindent \textbf{Reinforcement Learning for Multi-Agent Orchestration.}
Reinforcement learning provides a natural mechanism for optimizing LLM-based MAS from execution feedback. Existing methods train policies with rewards defined over final task success, execution efficiency, preference alignment, model or tool usage, and topology complexity \cite{nielsen2025learning, su2025toolorchestra, gao2025flowreasoner, wangtopoweaver, jiang2026fd}. Recent topology-focused work further uses group-relative optimization to reduce reward noise and assign finer-grained credit to communication edges \cite{cang2026graph}. Execution-level rewards alone do not identify the contribution of individual orchestration decisions. Our approach obtains a field-specific training signal by comparing specifications that differ in one duty, capacity, or dependency field, and applying the resulting reward contrast to the corresponding token spans.
% 没有太多related work了 
\section{Method}
\begin{figure}
    \centering
    \includegraphics[width=1\linewidth]{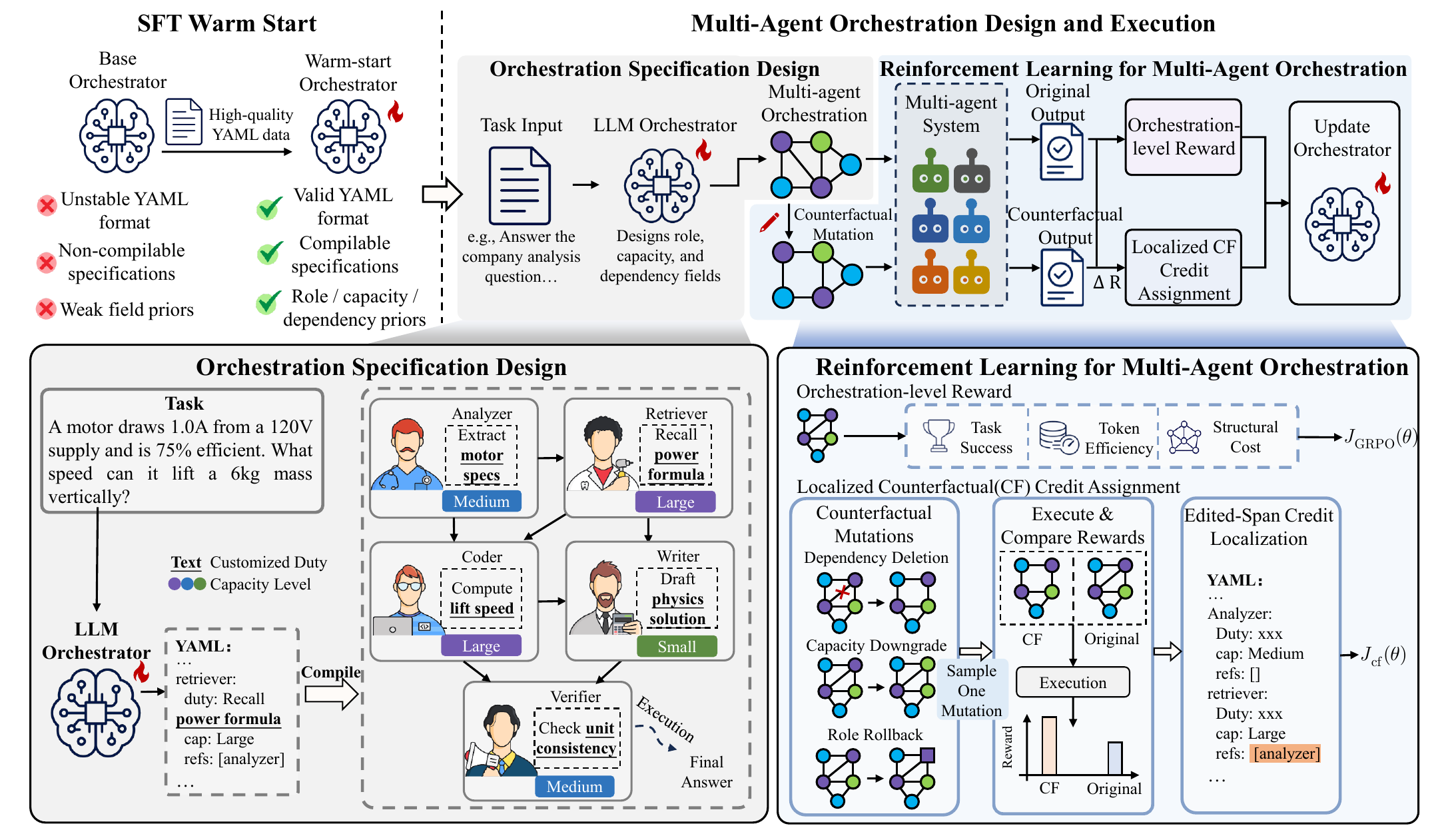}
    \caption{Overview of \methodname{}. The orchestrator designs a unified orchestration specification over roles, capacities, and dependencies. Training performs orchestration-level GRPO and localizes counterfactual reward contrasts to edited spans.}
    \label{fig:main_pipeline}
    \vspace{-8pt}
\end{figure}
We present \methodname, an LLM-based framework for learning compositional and efficient multi-agent orchestration design. Given an input task, an LLM orchestrator designs a structured orchestration specification consisting of agent entries, where each entry specifies a task-specific role, an assigned capacity level, and dependency references to other agents. The specification is written as a YAML object; after validation against an execution schema, it is compiled into an orchestration graph and executed by worker agents. Training combines an orchestration-level reward computed after executing the generated specification with a localized counterfactual comparison over a locally edited specification. The orchestration reward evaluates the generated design as a whole, while the counterfactual comparison edits one design field, namely role, capacity, or dependency, and attributes the resulting reward contrast to the edited span. Together, these signals optimize the specification as a complete orchestration design while providing localized credit for individual design choices. Figure~\ref{fig:main_pipeline} illustrates the overall pipeline.

\subsection{Problem Definition}
Let $x\in\mathcal{X}$ be a task instance and let $\pi_{\theta}$ denote the orchestrator. 
Given $x$, the orchestrator generates a unified multi-agent specification
\begin{equation}
    y\sim\pi_{\theta}(\cdot\mid x),
    \qquad
    y\in\mathcal{Y},
\end{equation}
where $\mathcal{Y}$ denotes the specification space. The generated specification $y$ defines how a multi-agent system is instantiated and executed for the input task.

Executing a valid specification $y$ produces a final answer $\hat{a}_y$ and an orchestration-level reward $R_{\mathrm{orch}}(x,y)$. Our goal is to learn an orchestrator that generates task-specific multi-agent specifications with high orchestration reward. We next introduce the orchestration design space and the reinforcement learning objective used to optimize $\pi_{\theta}$.

\subsection{Multi-agent Orchestration Design}
\label{sec:orchestration_design}

Given a task $x$, the orchestrator jointly specifies what each agent does, which worker capacity it uses, and which upstream outputs it receives. These choices are interdependent: the capacity needed to fulfill an agent's duty depends on both the assigned work and the information available from its dependencies. We therefore represent them within a single specification $y$ generated by $\pi_\theta$. We next formalize the resulting orchestration graph, describe how these components are represented in YAML, and introduce the rule-based validation applied before execution. Orchestration execution and learning are presented in Section~\ref{sec:rl}.

\noindent\textbf{Orchestration specification.}
For a valid specification $y$, the resulting multi-agent system is represented as a directed graph
\begin{equation}
    G_y=(V_y,E_y),
\end{equation}
where $V_y$ is the set of instantiated agents and $E_y\subseteq V_y\times V_y$ is the set of dependency relations. Each agent $v\in V_y$ is specified by
\begin{equation}
    v=(b_v,d_v,c_v),
    \qquad
    c_v\in\mathcal{L},
\end{equation}
where $b_v$ denotes its base role, $d_v$ denotes its task-specific duty, and $c_v$ denotes its assigned capacity level from the available worker capacities $\mathcal{L}$. Each dependency $(u,v)\in E_y$ indicates that the output of agent $u$ is provided as input context to agent $v$.

\noindent\textbf{YAML specification.}
The specification $y$ is written as a YAML object consisting of agent entries and dependency references. Each agent entry records its base role, task-specific duty, and capacity level, together with its dependency references to other agents. Agents in the same YAML step can execute concurrently, while references to agents in preceding steps determine how intermediate outputs are propagated across the orchestration.

\noindent\textbf{Rule-based validation.}
Before execution, a rule-based parser checks whether the generated specification satisfies the YAML schema and execution constraints. Specifically, it verifies the required agent attributes, valid capacity values, unique agent identifiers, and references to agents in preceding steps. We denote the validation result by
\begin{equation}
    \mathrm{Valid}(y)\in\{0,1\}.
\end{equation}
Only specifications satisfying $\mathrm{Valid}(y)=1$ define $G_y$ and are passed to worker execution. Because dependency references can only target agents in preceding steps, the resulting orchestration graph is directed and acyclic. Appendix~\ref{app:yaml_example} provides the complete YAML schema, validation rules, and an example specification.

\subsection{Reinforcement Learning for Multi-Agent Orchestration}
\label{sec:rl}

We initialize the orchestrator with supervised fine-tuning on executable specifications, then optimize it using orchestration-level GRPO and localized counterfactual supervision. Details of supervised initialization are provided in Appendix~\ref{app:sft_warm_start}. The two objectives are combined to train the orchestrator using both complete-specification rewards and localized field comparisons.

\noindent\textbf{Orchestration execution.}
For a valid specification $y$, the worker agents execute $G_y$ according to the dependency relations in $E_y$. Each agent $v\in V_y$ produces
\begin{equation}
    o_v
    =
    f_{c_v}\!\left(
        x,
        b_v,
        d_v,
        \{o_u:(u,v)\in E_y\}
    \right),
\end{equation}
where $f_{c_v}$ is the worker model associated with capacity level $c_v$, and the outputs of agents on which $v$ depends are provided as input context. The designated terminal agent returns the final answer $\hat{a}_y$. Let $n_v^{\mathrm{in}}$ and $n_v^{\mathrm{out}}$ denote the input- and output-token counts of worker $v$, respectively. We use the same capacity-weighted worker-token cost for reward computation and evaluation:
\begin{equation}
    C(x,y)
    =
    \sum_{v\in V_y}
    \omega_{\mathrm{cap}}(c_v)
    \left(n_v^{\mathrm{in}}+n_v^{\mathrm{out}}\right),
    \label{eq:worker_cost}
\end{equation}
where $\omega_{\mathrm{cap}}(c_v)$ weights both input and output tokens by the assigned worker capacity. The capacity weights and evaluation aggregation are specified in Appendix~\ref{eval_metric}.

\noindent\textbf{Global orchestration reward. }
For a generated specification $y$, we define the orchestration reward as
\begin{equation}
R_{\mathrm{orch}}(x,y)
=
\begin{cases}
R_{\mathrm{task}}(x,\hat{a}_y)
+\lambda_{\mathrm{tok}}B_{\mathrm{tok}}(x,y)
-\lambda_{\mathrm{str}}\Omega_{\mathrm{str}}(G_y),
&
\mathrm{Valid}(y)=1,
\\[2pt]
r_{\mathrm{invalid}},
&
\mathrm{Valid}(y)=0,
\end{cases}
\label{eq:orch_reward}
\end{equation}
where $R_{\mathrm{task}}$ measures task success, $B_{\mathrm{tok}}$ is the token-efficiency bonus, $\Omega_{\mathrm{str}}$ is the structural cost, and $r_{\mathrm{invalid}}<0$ penalizes invalid specifications. The coefficients $\lambda_{\mathrm{tok}}$ and $\lambda_{\mathrm{str}}$ control the contributions of the efficiency bonus and structural cost, respectively. We define the structural cost as
\begin{equation}
    \Omega_{\mathrm{str}}(G_y)
    =
    \omega_V|V_y|
    +
    \omega_E|E_y|
    +
    \omega_Hh(G_y),
    \label{eq:structural_cost}
\end{equation}
where $h(G_y)$ is the depth of the orchestration graph, and $\omega_V$, $\omega_E$, and $\omega_H$ weight the costs of agents, dependencies, and graph depth.

To compute the token-efficiency bonus, we map the capacity-weighted worker-token cost $C(x,y)$ using task-specific lower and upper bounds, $\tau_x^{-}$ and $\tau_x^{+}$, with $\tau_x^{-}<\tau_x^{+}$. These bounds remain fixed during RL and define the cost interval over which the efficiency bonus decreases. Costs at or below the lower bound receive the maximum efficiency bonus, whereas costs at or above the upper bound receive no efficiency bonus. Within this interval, the bonus decreases linearly as worker-token usage increases. The bounded token-efficiency bonus is
\begin{equation}
B_{\mathrm{tok}}(x,y)
=
g_y
\left[
\frac{\tau_x^{+}-C(x,y)}
{\tau_x^{+}-\tau_x^{-}}
\right]_{0}^{1},
\qquad
g_y
=
\begin{cases}
1,
&
\hat{a}_y\text{ is correct},
\\
\rho,
&
y\text{ is valid but unsolved},
\end{cases}
\label{eq:token_bonus}
\end{equation}
where $[z]_{0}^{1}=\min(1,\max(0,z))$ and $\rho\in[0,1)$ is the reduced bonus coefficient for a valid but unsolved specification. Costs at or below $\tau_x^{-}$ receive the maximum efficiency bonus, whereas costs at or above $\tau_x^{+}$ receive no efficiency bonus. The reward weights are reported in Appendix~\ref{app:reward_hyperparams}.

For each training task $x\sim\mathcal{D}_{\mathrm{train}}$, we sample $K$ specifications from $\pi_{\theta_{\mathrm{old}}}(\cdot\mid x)$ and compute their rewards $R_i=R_{\mathrm{orch}}(x,y_i)$. We minimize the standard clipped GRPO loss~\cite{guo2025deepseek}:
\begin{equation}
J_{\mathrm{GRPO}}(\theta)
=
-\mathbb{E}_{x,y_{1:K}}\!\left[
\frac{1}{K}\sum_{i=1}^{K}
\frac{1}{|y_i|}\sum_{t=1}^{|y_i|}
\ell_{\mathrm{clip}}\!\left(
r_{i,t}(\theta),\widehat{A}_i
\right)
\right],
\label{eq:grpo_loss}
\end{equation}
where $\widehat{A}_i$ is the group-normalized reward, $r_{i,t}$ is the token-wise importance ratio, and $\ell_{\mathrm{clip}}$ is the clipped surrogate defined in Appendix~\ref{app:grpo_objective}. GRPO assigns the same advantage to all tokens in a specification, providing a global signal without distinguishing individual design fields.

\noindent\textbf{Localized counterfactual credit assignment.}
To obtain credit for an individual orchestration field, we compare a factual specification $y$ with a counterfactual specification that differs in exactly one field. We consider the counterfactual families $\mathcal{F}_{\mathrm{cf}}=\{\varphi_{\mathrm{dep}},\varphi_{\mathrm{role}},\varphi_{\mathrm{cap}}\}$, corresponding to dependency deletion, role rollback, and capacity downgrade, respectively. Dependency deletion removes one feasible upstream reference, role rollback replaces $d_v$ with the generic duty associated with $b_v$, and capacity downgrade replaces $c_v$ with the next lower capacity level.

Given a feasible counterfactual family $\varphi\in\mathcal{F}_{\mathrm{cf}}$ and a valid field location $\ell$, the edit operator constructs $\widetilde{y}=\mu_{\varphi,\ell}(y)$. Sampling the counterfactual family and field location induces the conditional distribution $q_{\mathrm{cf}}(\cdot\mid y)$, from which $\widetilde{y}\sim q_{\mathrm{cf}}(\cdot\mid y)$. Let $S^{\mathrm{fact}}$ and $S^{\mathrm{cf}}$ denote the token spans of the selected field values in $y$ and $\widetilde{y}$, respectively. Mutation sampling and span construction are described in Appendix~\ref{app:cf_hyperparams}.

We evaluate $y$ and $\widetilde{y}$ using the same orchestration reward and compute
\begin{equation}
    \Delta R
    =
    R_{\mathrm{orch}}(x,y)
    -
    R_{\mathrm{orch}}(x,\widetilde{y}).
    \label{eq:cf_delta}
\end{equation}
The sign of $\Delta R$ indicates whether the factual or counterfactual field value produces the higher orchestration reward. Because the two specifications differ in only one field, their reward contrast provides localized credit for that field.

We use this contrast to favor the field value with the higher orchestration reward, applying the supervision only to the selected field-value spans. For a serialized specification $z$ and a nonempty field span $U$, we define the average log-probability
\begin{equation}
    s_{\theta}(x,z,U)
    =
    \frac{1}{|U|}
    \sum_{t\in U}
    \log\pi_{\theta}(z_t\mid x,z_{<t}).
\end{equation}
We then set $s_{\theta}^{\mathrm{fact}}=s_{\theta}(x,y,S^{\mathrm{fact}})$ and $s_{\theta}^{\mathrm{cf}}=s_{\theta}(x,\widetilde{y},S^{\mathrm{cf}})$. Tokens outside these spans remain conditioning context but do not contribute directly to the localized loss. We use the contrast weight
\begin{equation}
    w_{\mathrm{cf}}(\Delta R)=
    \frac{\min(|\Delta R|,\delta_{\max})}{\delta_{\max}},
    \qquad \delta_{\max}>0,
    \label{eq:cf_weight}
\end{equation}
where $\delta_{\max}$ limits the influence of large reward contrasts. The localized counterfactual loss is
\begin{equation}
\begin{aligned}
J_{\mathrm{cf}}(\theta)
={}&-\mathbb{E}_{x,y,\widetilde{y}}
\Bigl[\mathbf{1}\{|\Delta R|\geq\delta_{\min}\}\,w_{\mathrm{cf}}(\Delta R)\\[-2pt]
&\qquad\cdot\log\operatorname{sigmoid}\!\left(
\beta_{\mathrm{cf}}\operatorname{sign}(\Delta R)
\left[s_{\theta}^{\mathrm{fact}}-s_{\theta}^{\mathrm{cf}}\right]
\right)\Bigr].
\end{aligned}
\label{eq:cf_loss}
\end{equation}
Here, $\operatorname{sigmoid}(u)=(1+e^{-u})^{-1}$, $\beta_{\mathrm{cf}}>0$ controls comparison sharpness, and $\delta_{\min}>0$ filters negligible reward differences. The expectation is over valid sampled specifications and their feasible counterfactuals. Hyperparameter values are reported in Appendix~\ref{app:cf_hyperparams}. A theoretical analysis of the localized counterfactual objective is provided in Appendix~\ref{app:cf_local_analysis}.

\noindent\textbf{Overall optimization.}
We minimize the overall training loss
\begin{equation}
    J_{\mathrm{total}}(\theta)
    =
    J_{\mathrm{GRPO}}(\theta)
    +
    \lambda_{\mathrm{cf}}
    J_{\mathrm{cf}}(\theta),
    \label{eq:total_loss}
\end{equation}
where $\lambda_{\mathrm{cf}}$ controls the contribution of localized counterfactual credit assignment. The GRPO term optimizes complete orchestration specifications, whereas the counterfactual term provides localized supervision for the tested field. Counterfactual execution is used only during training, and unchanged worker calls can be reused through the execution cache described in Appendix~\ref{app:cf_overhead}.
\section{Experiments}\label{sec:experiment}
We evaluate \methodname{} across reasoning and coding tasks. We compare task performance against single-agent prompting, fixed-structure MAS, topology-design, and adaptive workflow baselines, and further analyze the accuracy--token trade-off of the learned orchestration. We also ablate supervised initialization, orchestration-level GRPO, role--capacity--dependency design components, and localized counterfactual mutation families, followed by case studies.

\subsection{Experimental Setup}
\noindent\textbf{Benchmarks.}
We evaluate on six benchmarks: GSM8K~\cite{cobbe2021training}, MultiArith~\cite{roy2015solving}, SVAMP~\cite{patel2021nlp}, and AQuA~\cite{ling2017program} for mathematical reasoning; MMLU~\cite{hendrycks2020measuring} for multiple-choice reasoning; and HumanEval~\cite{chen2021evaluating} for code generation. For each task, the orchestrator generates a specification that is validated, compiled, and executed to produce the final answer. Evaluation metrics and dataset splits are detailed in Appendices~\ref{eval_metric} and~\ref{app:evaluation_protocol}.

\noindent\textbf{Baselines.}
We compare against four groups of baselines: single-agent prompting with Vanilla, CoT~\cite{wei2022chain}, and SC-CoT~\cite{wang2022self}; fixed-structure MAS with Chain, Tree, Complete, and Random topologies; topology-design methods including AgentPrune~\cite{zhang2024cut}, AgentDropout~\cite{wang2025agentdropout}, G-Designer~\cite{zhang2024g}, ARG-Designer~\cite{li2026assemble}, and OFA-MAS~\cite{li2026ofa}; and the adaptive workflow method AFlow~\cite{zhang2024aflow}.

\mainresultstable

\noindent\textbf{Performance.}
Table~\ref{tab:main_results} compares performance across six benchmarks. \methodname{} achieves the highest average score of 90.72, outperforming SC-CoT, Tree-MAS, AgentPrune, and AFlow by 3.96, 2.19, 2.59, and 5.39 percentage points, respectively. It ranks first on five benchmarks and second on MMLU, where it trails ARG-Designer by only 0.98 points. These results demonstrate the effectiveness of jointly designing task-specific agent duties, capacity levels, and dependency relations within a unified orchestration specification.

\begin{figure}
    \centering
    \includegraphics[width=1\linewidth]{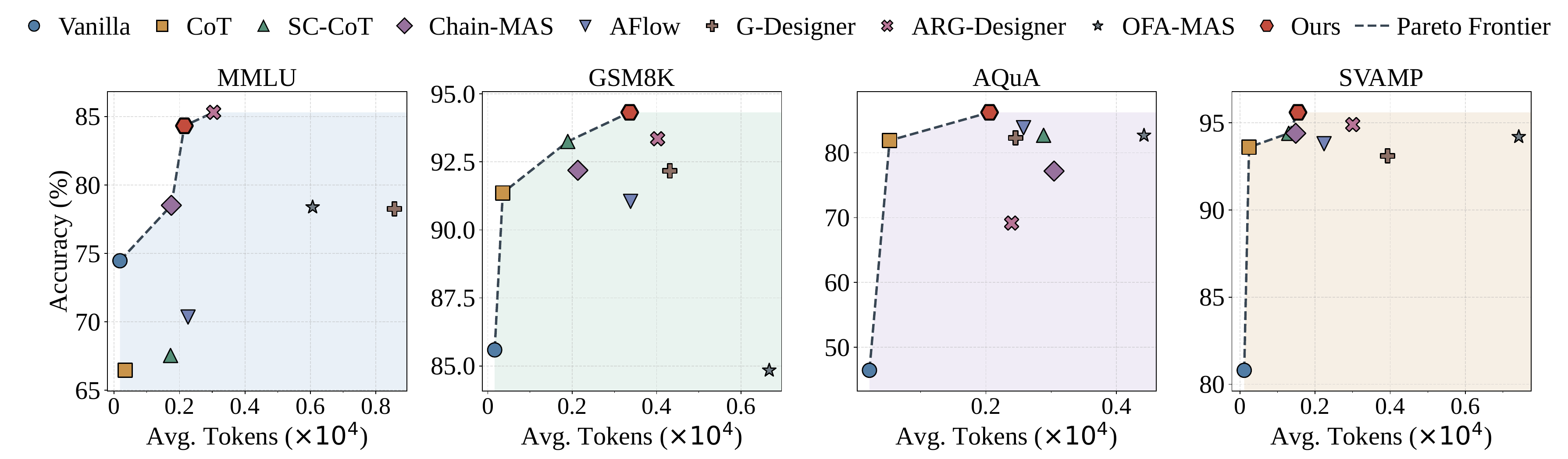}
\caption{Accuracy--token trade-off on four benchmarks. Token usage is the average capacity-weighted sum of worker input and output tokens per task. Dashed lines connect methods on the Pareto frontier.}
    \label{fig:token_efficiency}
    \vspace{-12pt}
\end{figure}

\noindent\textbf{Token Efficiency.}
Figure~\ref{fig:token_efficiency} compares task accuracy against capacity-weighted worker-token usage, using the cost $C(x,y)$ defined in Section~\ref{sec:rl} and averaged over evaluation tasks. \methodname{} lies on the Pareto frontier on all four benchmarks. On MMLU, it achieves competitive accuracy using approximately $0.2\times10^4$ worker tokens, compared with $0.3\times10^4$ for ARG-Designer and more than $0.6\times10^4$ for several topology-design baselines. On GSM8K, AQuA, and SVAMP, it achieves the highest accuracy while using fewer worker tokens than several topology-design baselines. Together, these results show that task-adaptive capacity assignment and dependency construction improve the accuracy--token trade-off.

\ablationtable

\noindent\textbf{Ablation Study.}
Table~\ref{tab:ablation} examines the training stages and orchestration design components on MMLU, GSM8K, and AQuA. SFT substantially improves specification validity, but SFT only remains below the strongest single-agent result on each task (Appendix~\ref{app:training_stages}). Starting from the same SFT checkpoint, orchestration-level GRPO improves accuracy by 10.19 / 11.75 / 5.90 percentage points. Localized counterfactual credit assignment adds a further 1.24 / 0.92 / 0.79 points while reducing worker tokens by 15.5\% / 10.2\% / 31.7\%. These results show that supervised initialization provides a foundation for executable generation, while execution-based RL learns more effective orchestration and counterfactual supervision further improves both accuracy and efficiency.

Removing any orchestration design component reduces accuracy. Dependency construction has the largest effect on MMLU (-3.46 points), while role design contributes most on AQuA (-3.15 points), highlighting the task-dependent importance of information flow and specialized duties. Capacity assignment improves both accuracy and efficiency: removing it more than doubles worker-token usage on AQuA, from 2056 to 4154, while reducing accuracy. Removing any counterfactual mutation family also lowers accuracy, supporting the complementary contributions of dependency-, role-, and capacity-level credit assignment.

\begin{wrapfigure}{r}{0.55\linewidth}
    \vspace{-12pt}
    \centering
    \includegraphics[width=\linewidth]{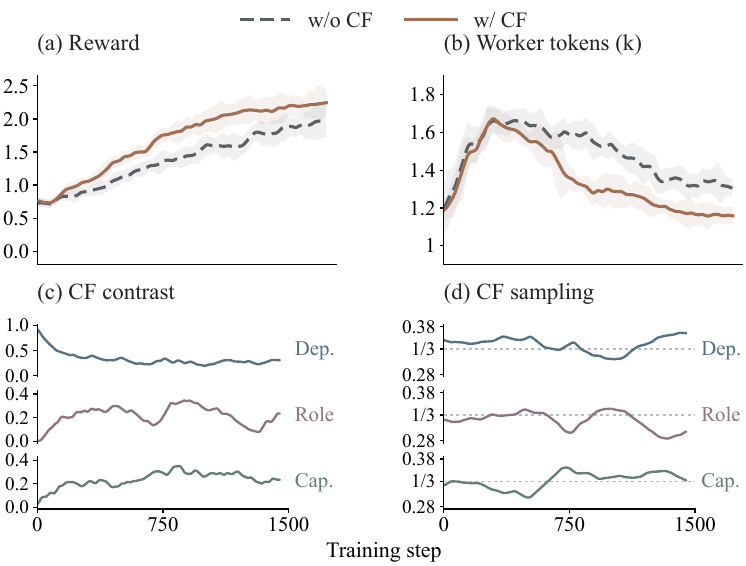}
    \caption{Training dynamics with and without localized counterfactual credit assignment in (a) orchestration reward and (b) worker tokens per execution, in thousands. For the full method, (c) shows absolute reward contrasts and (d) shows adaptive sampling probabilities for dependency deletion (Dep.), role rollback (Role), and capacity downgrade (Cap.).}
    \label{fig:cf_analyse}
    \vspace{-8pt}
\end{wrapfigure}

\noindent\textbf{Training Dynamics.}
Figure~\ref{fig:cf_analyse} shows the effect of localized counterfactual credit assignment on training. Shaded regions in panels (a) and (b) indicate $\pm 1$ standard deviation across random seeds. Under otherwise identical training settings, the full method increases the final reward from approximately 2.0 to 2.25 compared with orchestration-level GRPO alone. This improvement involves additional counterfactual executions during training, whose overhead is reduced by node-level caching (Appendix~\ref{app:cf_overhead}). In return, final worker-token usage decreases by approximately 11.1\%, from 1.30K to 1.15K per execution. Panel (c) shows the absolute reward contrast $|\Delta R|$ for each mutation family: larger values indicate a stronger effect of the tested edits on orchestration reward. Dependency deletion yields the largest early contrasts, while role rollback and capacity downgrade provide sustained signals later in training. Panel (d) shows each mutation family's sampling probability, with the dashed $1/3$ line indicating uniform sampling. The probabilities vary around this reference, allocating more comparisons to families with larger recent reward contrasts while maintaining exploration across all three families.

\begin{figure}
    \centering
    \includegraphics[width=1\linewidth]{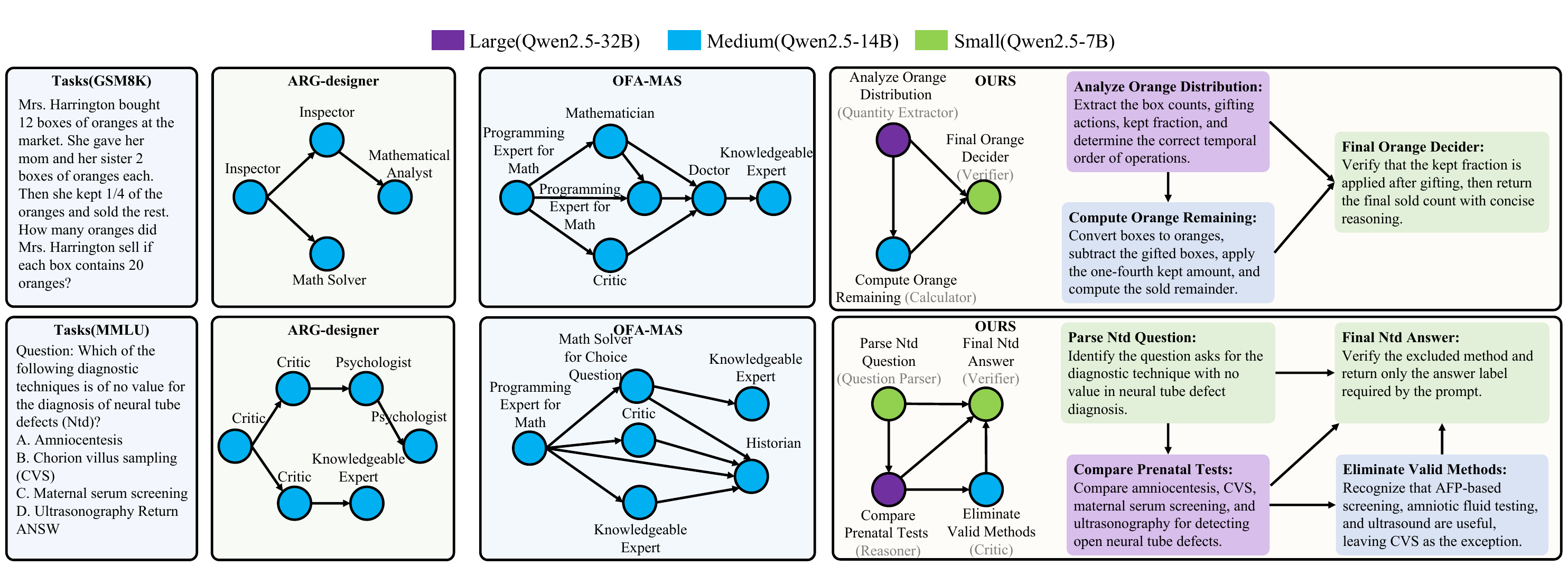}
    \caption{Case studies on GSM8K and MMLU. Node colors denote capacity levels, inherited base roles are shown as \textcolor{gray!65}{(Base Role)}, and the expanded views show \methodname{}'s customized duties and dependency references.}
    \label{fig:case_study}
    \vspace{-12pt}
\end{figure}

\noindent\textbf{Case Study.}
Figure~\ref{fig:case_study} provides qualitative examples on GSM8K and MMLU. Compared with ARG-Designer and OFA-MAS, \methodname{} designs more task-aligned orchestration specifications rather than only adapting the graph topology. In the GSM8K example, the generated agents correspond to the main reasoning steps: analyzing the orange distribution, computing the remaining oranges, and verifying the final answer. In the MMLU example, \methodname{} decomposes the medical question into parsing the neural tube defect diagnosis target, comparing prenatal tests, identifying the exception among valid diagnostic methods, and producing the final answer. The expanded views show that customized duties and dependency references follow these reasoning steps, while capacity levels are assigned according to each agent's expected workload. In contrast, ARG-Designer and OFA-MAS use broader or repeated roles in these examples, such as \textit{Inspector}, \textit{Critic}, and \textit{Knowledgeable Expert}, whose responsibilities are less directly tied to the task requirements. These cases illustrate the joint adaptation of role specifications, capacity levels, and dependency references to task-specific reasoning.
\section{Conclusion}
We presented \methodname{}, a framework for learning compositional and efficient multi-agent orchestration. It jointly designs task-specific role specifications, capacity levels, and dependency references within a unified executable orchestration specification, adapting the design to each task. Training combines orchestration-level GRPO with localized counterfactual credit assignment, translating orchestration-reward contrasts into span-level supervision for edited fields. This allows the orchestrator to refine individual orchestration fields while optimizing complete specifications. Experiments across reasoning and coding benchmarks demonstrate strong task performance and token efficiency compared with single-agent prompting and existing multi-agent orchestration methods. Ablations and training-dynamics analyses further highlight the complementary roles of the orchestration design components and the benefits of localized counterfactual credit assignment.

\subsection*{AI use statement}
We used generative AI tools for language polishing, grammar correction, and improving the clarity of the manuscript, as well as for LaTeX formatting and consistency checks of notation and reported numerical summaries. The authors reviewed the AI-assisted revisions and take full responsibility for the accuracy and integrity of the final manuscript.

\subsection*{Reproducibility statement}
To support reproducibility, we provide the orchestration schema
and an example specification in
Appendix~\ref{app:yaml_example}, evaluation metrics in
Appendix~\ref{eval_metric}, and prompt templates in
Appendix~\ref{app:prompts}.
Appendix~\ref{app:reproducibility} documents the model deployment,
training hyperparameters, supervised warm start, and evaluation
protocol, including dataset splits and evaluation sizes.
The assumptions and derivation underlying our local
counterfactual analysis are provided in
Appendix~\ref{app:cf_local_analysis}.
Execution caching and computational resources are described in
Appendices~\ref{app:cf_overhead} and~\ref{app:compute}, respectively.
An anonymous code repository is linked in the abstract.

\newpage
\bibliographystyle{iclr2027_conference}
\bibliography{iclr2027_conference}

@inproceedings{wu2024autogen,
  title={Autogen: Enabling next-gen LLM applications via multi-agent conversations},
  author={Wu, Qingyun and Bansal, Gagan and Zhang, Jieyu and Wu, Yiran and Li, Beibin and Zhu, Erkang and Jiang, Li and Zhang, Xiaoyun and Zhang, Shaokun and Liu, Jiale and others},
  booktitle={First conference on language modeling},
  year={2024}
}

@article{li2023camel,
  title={Camel: Communicative agents for" mind" exploration of large language model society},
  author={Li, Guohao and Hammoud, Hasan and Itani, Hani and Khizbullin, Dmitrii and Ghanem, Bernard},
  journal={Advances in neural information processing systems},
  volume={36},
  pages={51991--52008},
  year={2023}
}

@inproceedings{hong2023metagpt,
  title={MetaGPT: Meta programming for a multi-agent collaborative framework},
  author={Hong, Sirui and Zhuge, Mingchen and Chen, Jonathan and Zheng, Xiawu and Cheng, Yuheng and Wang, Jinlin and Zhang, Ceyao and Wang, Zili and Yau, Steven Ka Shing and Lin, Zijuan and others},
  booktitle={The twelfth international conference on learning representations},
  year={2023}
}

@article{guo2024large,
  title={Large language model based multi-agents: A survey of progress and challenges},
  author={Guo, Taicheng and Chen, Xiuying and Wang, Yaqi and Chang, Ruidi and Pei, Shichao and Chawla, Nitesh V and Wiest, Olaf and Zhang, Xiangliang},
  journal={arXiv preprint arXiv:2402.01680},
  year={2024}
}

@inproceedings{du2024improving,
  title={Improving factuality and reasoning in language models through multiagent debate},
  author={Du, Yilun and Li, Shuang and Torralba, Antonio and Tenenbaum, Joshua B and Mordatch, Igor},
  booktitle={Forty-first international conference on machine learning},
  year={2024}
}

@article{yao2023tree,
  title={Tree of thoughts: Deliberate problem solving with large language models},
  author={Yao, Shunyu and Yu, Dian and Zhao, Jeffrey and Shafran, Izhak and Griffiths, Tom and Cao, Yuan and Narasimhan, Karthik},
  journal={Advances in neural information processing systems},
  volume={36},
  pages={11809--11822},
  year={2023}
}

@inproceedings{zhuge2024gptswarm,
  title={Gptswarm: Language agents as optimizable graphs},
  author={Zhuge, Mingchen and Wang, Wenyi and Kirsch, Louis and Faccio, Francesco and Khizbullin, Dmitrii and Schmidhuber, J{\"u}rgen},
  booktitle={Forty-first International Conference on Machine Learning},
  year={2024}
}

@article{zhang2024g,
  title={G-designer: Architecting multi-agent communication topologies via graph neural networks},
  author={Zhang, Guibin and Yue, Yanwei and Sun, Xiangguo and Wan, Guancheng and Yu, Miao and Fang, Junfeng and Wang, Kun and Chen, Tianlong and Cheng, Dawei},
  journal={arXiv preprint arXiv:2410.11782},
  year={2024}
}

@article{zhang2024cut,
  title={Cut the crap: An economical communication pipeline for llm-based multi-agent systems},
  author={Zhang, Guibin and Yue, Yanwei and Li, Zhixun and Yun, Sukwon and Wan, Guancheng and Wang, Kun and Cheng, Dawei and Yu, Jeffrey Xu and Chen, Tianlong},
  journal={arXiv preprint arXiv:2410.02506},
  year={2024}
}

@article{zhang2024aflow,
  title={Aflow: Automating agentic workflow generation},
  author={Zhang, Jiayi and Xiang, Jinyu and Yu, Zhaoyang and Teng, Fengwei and Chen, Xionghui and Chen, Jiaqi and Zhuge, Mingchen and Cheng, Xin and Hong, Sirui and Wang, Jinlin and others},
  journal={arXiv preprint arXiv:2410.10762},
  year={2024}
}

@article{fan2024workflowllm,
  title={Workflowllm: Enhancing workflow orchestration capability of large language models},
  author={Fan, Shengda and Cong, Xin and Fu, Yuepeng and Zhang, Zhong and Zhang, Shuyan and Liu, Yuanwei and Wu, Yesai and Lin, Yankai and Liu, Zhiyuan and Sun, Maosong},
  journal={arXiv preprint arXiv:2411.05451},
  year={2024}
}

@article{cai2025agentbalance,
  title={AgentBalance: Backbone-then-Topology Design for Cost-Effective Multi-Agent Systems under Budget Constraints},
  author={Cai, Shuowei and Ning, Yansong and Liu, Hao},
  journal={arXiv preprint arXiv:2512.11426},
  year={2025}
}

@inproceedings{li2026assemble,
  title={Assemble your crew: Automatic multi-agent communication topology design via autoregressive graph generation},
  author={Li, Shiyuan and Liu, Yixin and Wen, Qingsong and Zhang, Chengqi and Pan, Shirui},
  booktitle={Proceedings of the AAAI Conference on Artificial Intelligence},
  volume={40},
  number={28},
  pages={23142--23150},
  year={2026}
}

@inproceedings{li2026ofa,
  title={OFA-MAS: One-for-All Multi-Agent System Topology Design based on Mixture-of-Experts Graph Generative Models},
  author={Li, Shiyuan and Liu, Yixin and Zheng, Yu and Li, Mei and Nguyen, Quoc Viet Hung and Pan, Shirui},
  booktitle={Proceedings of the ACM Web Conference 2026},
  pages={1333--1344},
  year={2026}
}

@article{hu2024automated,
  title={Automated design of agentic systems},
  author={Hu, Shengran and Lu, Cong and Clune, Jeff},
  journal={arXiv preprint arXiv:2408.08435},
  year={2024}
}

@article{nielsen2025learning,
  title={Learning to Orchestrate Agents in Natural Language with the Conductor},
  author={Nielsen, Stefan and Cetin, Edoardo and Schwendeman, Peter and Sun, Qi and Xu, Jinglue and Tang, Yujin},
  journal={arXiv preprint arXiv:2512.04388},
  year={2025}
}

@inproceedings{su2026difficulty,
  title={Difficulty-aware agentic orchestration for query-specific multi-agent workflows},
  author={Su, Jinwei and Lan, Qizhen and Xia, Yinghui and Sun, Lifan and Tian, Weiyou and Shi, Tianyu and He, Lewei},
  booktitle={Proceedings of the ACM Web Conference 2026},
  pages={2060--2070},
  year={2026}
}

@article{zhang2025multi,
  title={Multi-agent architecture search via agentic supernet},
  author={Zhang, Guibin and Niu, Luyang and Fang, Junfeng and Wang, Kun and Bai, Lei and Wang, Xiang},
  journal={arXiv preprint arXiv:2502.04180},
  year={2025}
}

@inproceedings{yue2025masrouter,
  title={Masrouter: Learning to route llms for multi-agent systems},
  author={Yue, Yanwei and Zhang, Guibin and Liu, Boyang and Wan, Guancheng and Wang, Kun and Cheng, Dawei and Qi, Yiyan},
  booktitle={Proceedings of the 63rd Annual Meeting of the Association for Computational Linguistics (Volume 1: Long Papers)},
  pages={15549--15572},
  year={2025}
}

@article{yao2026hieramas,
  title={HieraMAS: Optimizing Intra-Node LLM Mixtures and Inter-Node Topology for Multi-Agent Systems},
  author={Yao, Tianjun and Li, Zhaoyi and Shen, Zhiqiang},
  journal={arXiv preprint arXiv:2602.20229},
  year={2026}
}

@article{su2025toolorchestra,
  title={Toolorchestra: Elevating intelligence via efficient model and tool orchestration},
  author={Su, Hongjin and Diao, Shizhe and Lu, Ximing and Liu, Mingjie and Xu, Jiacheng and Dong, Xin and Fu, Yonggan and Belcak, Peter and Ye, Hanrong and Yin, Hongxu and others},
  journal={arXiv preprint arXiv:2511.21689},
  year={2025}
}

@article{gao2025flowreasoner,
  title={Flowreasoner: Reinforcing query-level meta-agents},
  author={Gao, Hongcheng and Liu, Yue and He, Yufei and Dou, Longxu and Du, Chao and Deng, Zhijie and Hooi, Bryan and Lin, Min and Pang, Tianyu},
  journal={arXiv preprint arXiv:2504.15257},
  year={2025}
}

@article{wangtopoweaver,
  title={TopoWeaver-R1: Reinforcing Difficulty-Aware Topology Evolution in Multi-Agent Competition-Level Code Generation},
  author={Wang, Siyu and Lu, Ruotian and Yang, Zhihao and Wang, Yuchao and Xu, Lei and Xu, Qimin and Yin, Guojun and Chen, Cailian and Guan, Xinping and others}
}

@inproceedings{jiang2026fd,
  title={FD-MAGRPO: Functionality-Driven Multi-Agent Group Relative Policy Optimization for Analog-LDO Sizing},
  author={Jiang, Haoning and Wu, Han and Ouyang, Zhuoli and Wang, Ziheng and Chen, Tinghuan and Jiang, Junmin},
  booktitle={Proceedings of the AAAI Conference on Artificial Intelligence},
  volume={40},
  number={27},
  pages={22310--22317},
  year={2026}
}

@article{cobbe2021training,
  title={Training verifiers to solve math word problems},
  author={Cobbe, Karl and Kosaraju, Vineet and Bavarian, Mohammad and Chen, Mark and Jun, Heewoo and Kaiser, Lukasz and Plappert, Matthias and Tworek, Jerry and Hilton, Jacob and Nakano, Reiichiro and others},
  journal={arXiv preprint arXiv:2110.14168},
  year={2021}
}

@inproceedings{patel2021nlp,
  title={Are NLP models really able to solve simple math word problems?},
  author={Patel, Arkil and Bhattamishra, Satwik and Goyal, Navin},
  booktitle={Proceedings of the 2021 conference of the North American chapter of the association for computational linguistics: human language technologies},
  pages={2080--2094},
  year={2021}
}

@inproceedings{ling2017program,
  title={Program induction by rationale generation: Learning to solve and explain algebraic word problems},
  author={Ling, Wang and Yogatama, Dani and Dyer, Chris and Blunsom, Phil},
  booktitle={Proceedings of the 55th annual meeting of the association for computational linguistics (volume 1: Long papers)},
  pages={158--167},
  year={2017}
}

@article{hendrycks2020measuring,
  title={Measuring massive multitask language understanding},
  author={Hendrycks, Dan and Burns, Collin and Basart, Steven and Zou, Andy and Mazeika, Mantas and Song, Dawn and Steinhardt, Jacob},
  journal={arXiv preprint arXiv:2009.03300},
  year={2020}
}

@article{chen2021evaluating,
  title={Evaluating large language models trained on code},
  author={Chen, Mark and Tworek, Jerry and Jun, Heewoo and Yuan, Qiming and Pinto, Henrique Ponde De Oliveira and Kaplan, Jared and Edwards, Harri and Burda, Yuri and Joseph, Nicholas and Brockman, Greg and others},
  journal={arXiv preprint arXiv:2107.03374},
  year={2021}
}

@inproceedings{roy2015solving,
  title={Solving general arithmetic word problems},
  author={Roy, Subhro and Roth, Dan},
  booktitle={Proceedings of the 2015 conference on empirical methods in natural language processing},
  pages={1743--1752},
  year={2015}
}

@inproceedings{wang2025agentdropout,
  title={Agentdropout: Dynamic agent elimination for token-efficient and high-performance llm-based multi-agent collaboration},
  author={Wang, Zhexuan and Wang, Yutong and Liu, Xuebo and Ding, Liang and Zhang, Miao and Liu, Jie and Zhang, Min},
  booktitle={Proceedings of the 63rd Annual Meeting of the Association for Computational Linguistics (Volume 1: Long Papers)},
  pages={24013--24035},
  year={2025}
}

@article{wei2022chain,
  title={Chain-of-thought prompting elicits reasoning in large language models},
  author={Wei, Jason and Wang, Xuezhi and Schuurmans, Dale and Bosma, Maarten and Xia, Fei and Chi, Ed and Le, Quoc V and Zhou, Denny and others},
  journal={Advances in neural information processing systems},
  volume={35},
  pages={24824--24837},
  year={2022}
}

@article{wang2022self,
  title={Self-consistency improves chain of thought reasoning in language models},
  author={Wang, Xuezhi and Wei, Jason and Schuurmans, Dale and Le, Quoc and Chi, Ed and Narang, Sharan and Chowdhery, Aakanksha and Zhou, Denny},
  journal={arXiv preprint arXiv:2203.11171},
  year={2022}
}

@inproceedings{cang2026graph,
  title={Graph-GRPO: Stabilizing Multi-Agent Topology Learning via Group Relative Policy Optimization},
  author={Cang, Yueyang and Zhang, Xiaoteng and Zhao, Erlu and Ji, Zehua and Liu, Yuhang and He, Yuchen and Ning, Zhiyuan and Yijun, Chen and Que, Wenge and Shi, Li},
  booktitle={Findings of the Association for Computational Linguistics: ACL 2026},
  pages={20222--20231},
  year={2026}
}

@article{yao2022react,
  title={React: Synergizing reasoning and acting in language models},
  author={Yao, Shunyu and Zhao, Jeffrey and Yu, Dian and Du, Nan and Shafran, Izhak and Narasimhan, Karthik and Cao, Yuan},
  journal={arXiv preprint arXiv:2210.03629},
  year={2022}
}

@article{shinn2023reflexion,
  title={Reflexion: Language agents with verbal reinforcement learning},
  author={Shinn, Noah and Cassano, Federico and Gopinath, Ashwin and Narasimhan, Karthik and Yao, Shunyu},
  journal={Advances in neural information processing systems},
  volume={36},
  pages={8634--8652},
  year={2023}
}

@inproceedings{qian2025scaling,
  title={Scaling large language model-based multi-agent collaboration},
  author={Qian, Chen and Xie, Zihao and Wang, Yifei and Liu, Wei and Zhu, Kunlun and Xia, Hanchen and Dang, Yufan and Du, Zhuoyun and Chen, Weize and Yang, Cheng and others},
  booktitle={International Conference on Learning Representations},
  volume={2025},
  pages={41488--41505},
  year={2025}
}

@article{guo2025deepseek,
  title={Deepseek-r1: Incentivizing reasoning capability in llms via reinforcement learning},
  author={Guo, Daya and Yang, Dejian and Zhang, Haowei and Song, Junxiao and Wang, Peiyi and Zhu, Qihao and Xu, Runxin and Zhang, Ruoyu and Ma, Shirong and Bi, Xiao and others},
  journal={arXiv preprint arXiv:2501.12948},
  year={2025}
}

\newpage
\appendix
\section{Orchestration Specification and Validation}
\label{app:yaml_example}
\label{app:yaml_spec}

An orchestration specification contains an optional \texttt{defaults} block and an ordered list of \texttt{steps}, each containing agent entries. Each active agent has a unique task-specific identifier \texttt{type}, an inherited \texttt{base\_role}, a customized \texttt{duty}, and dependency references \texttt{ref}. The runtime capacity is specified by \texttt{capacity\_level}, either explicitly or through the applicable step-level or default setting. Its allowed values are \texttt{small}, \texttt{medium}, and \texttt{large}; their worker backbones are listed in Appendix~\ref{app:model_deployment}.

The validator checks the schema, required fields, unique active identifiers, legal base roles and capacity levels, and a valid final-answer role. References in the first step must be empty; later references must point to active agents in preceding steps. These constraints ensure that a valid specification compiles into a layered directed acyclic graph. Agents in the same step can execute concurrently. The allowed base-role interfaces are listed in Appendix~\ref{app:role_interfaces}.

The example below illustrates the schema and information flow for a mathematical reasoning task. The duties are illustrative; in the experiments, the orchestrator generates task-specific duties for each input.

\begin{lstlisting}[
  basicstyle=\ttfamily\small,
  columns=fullflexible,
  keepspaces=true,
  showstringspaces=false,
  breaklines=true,
  breakatwhitespace=true,
  breakindent=1em,
  linewidth=\linewidth,
  xleftmargin=0pt,
  xrightmargin=0pt,
  frame=none
]
defaults:
  capacity_level: medium
steps:
  - agents:
      - type: extract_quantities
        base_role: math_analyst
        duty: Extract known quantities, unknown target, and explicit constraints.
        ref: []
        capacity_level: small

  - agents:
      - type: build_equations
        base_role: math_solver
        duty: Translate extracted quantities into arithmetic relations.
        ref: [extract_quantities]
        capacity_level: medium
      - type: check_units
        base_role: solution_inspector
        duty: Verify that each number is attached to the correct entity or unit.
        ref: [extract_quantities]
        capacity_level: small

  - agents:
      - type: compute_answer
        base_role: math_solver
        duty: Perform the arithmetic from the equation and unit check.
        ref: [build_equations, check_units]
        capacity_level: medium

  - agents:
      - type: verify_final_answer
        base_role: final_decider
        duty: Return the final answer after checking referenced arithmetic.
        ref: [compute_answer, check_units]
        capacity_level: medium
\end{lstlisting}

\lstdefinestyle{promptstyle}{
  basicstyle=\ttfamily\footnotesize,
  breaklines=true,
  breakatwhitespace=true,
  columns=fullflexible,
  keepspaces=true,
  showstringspaces=false,
  frame=none,
  xleftmargin=0pt,
  xrightmargin=0pt
}

\newtcblisting{promptbox}[1]{
  enhanced,
  breakable,
  listing only,
  listing engine=listings,
  listing options={style=promptstyle},
  colback=gray!3,
  colframe=gray!45,
  title={#1},
  fonttitle=\bfseries,
  boxrule=0.5pt,
  arc=2pt,
  left=4pt,
  right=4pt,
  top=4pt,
  bottom=4pt,
  boxsep=0pt,
  before skip=6pt,
  after skip=6pt
}

\section{Evaluation Metrics}\label{eval_metric}
For MMLU and AQuA, we report answer-option accuracy after deterministic answer extraction. For GSM8K, MultiArith, and SVAMP, we report exact-match accuracy after
normalizing the extracted numerical answer. For HumanEval, we report pass@1 under
unit-test execution. The average score in Table~\ref{tab:main_results} is the unweighted arithmetic mean of the six reported percentages. All task metrics are computed using deterministic evaluators and do not rely on an LLM judge.

Worker-token usage is the average capacity-weighted sum of worker input and output tokens per task, using the cost $C(x,y)$ defined in Section~\ref{sec:rl}. The capacity weights are
\begin{equation}
    \omega_{\mathrm{cap}}(c)=
    \begin{cases}
    0.5, & c=\mathrm{small}\ (\text{7B}),\\
    1, & c=\mathrm{medium}\ (\text{14B}),\\
    2, & c=\mathrm{large}\ (\text{32B}).
    \end{cases}
\end{equation}

For each evaluation instance $x$, let $y_x$ denote the executed specification. For methods without a generated YAML specification, $y_x$ denotes their instantiated worker configuration. The reported worker-token usage on evaluation dataset $\mathcal{D}_{\mathrm{eval}}$ is
\begin{equation}
    \overline{C}_{\mathcal{D}_{\mathrm{eval}}}
    =
    \frac{1}{|\mathcal{D}_{\mathrm{eval}}|}
    \sum_{x\in\mathcal{D}_{\mathrm{eval}}} C(x,y_x).
\end{equation}

\section{Prompts}
\label{app:prompts}

\subsection{Prompt Templates}
\label{app:prompt_templates}

We provide the prompt templates used in \methodname{} for orchestration specification design, YAML repair, and role-conditioned worker execution. Bracketed fields such as \texttt{[TASK]}, \texttt{[ALLOWED\_BASE\_ROLES]}, and \texttt{[REFERENCED\_OUTPUTS]} are filled by the runtime for each task instance.

The templates support configurable capacity-to-model mappings. In the reported experiments, \texttt{small}, \texttt{medium}, and \texttt{large} map to 7B, 14B, and 32B workers, respectively (Appendix~\ref{app:model_deployment}).

\begin{promptbox}{Orchestrator System Prompt}
You are a strict topology planner for a multi-agent reasoning system.
Return exactly one YAML code block and nothing else.
Generate one executable graph, inline custom roles, and runtime capacity_level allocation in a single pass.
Do not directly answer the task. Your output is only the orchestration specification.

Schema must be: optional defaults -> steps -> agents.

Hard constraints (must satisfy all):
1) Allowed top-level keys are only: defaults, steps.
2) Allowed nested keys are only: capacity_level, agents, type, base_role, duty, ref, active.
3) Do not introduce any other keys, comments, metadata fields, explanations, or natural-language notes.
4) Every active agent.type must be a unique custom role id for this topology.
5) The agent.type must not be a fixed-pool base role name; it should describe the task-specific local function of the agent.
6) Every active agent.base_role must come from Allowed base roles in the user prompt.
7) Every active agent.duty must be non-empty, problem-specific, and describe the local contribution expected from the agent.
8) Avoid generic duties such as "solve the problem" or "analyze the task" unless they specify a concrete local output.
9) Step-0 ref must be [].
10) Any ref item must reference an active agent.type that has already appeared in previous steps.
11) A ref item must not refer to the current agent, a later agent, an inactive agent, or an undefined agent.
12) The final active step must include a custom final-answer agent using one of the final base roles in the user prompt.
13) Valid capacity_levels are exactly: small, medium, large.
14) Include explicit runtime capacity_level information using defaults.capacity_level, step capacity_level, agent capacity_level, or a combination of them.
15) Use small for cheap extraction or formatting nodes, medium for normal reasoning nodes, and large only when stronger reasoning is clearly justified.
16) Keep the graph executable and concise. Do not add redundant agents, unnecessary steps, or avoidable dependency references.
17) The YAML must be parseable and must preserve the required hierarchy of steps and agents.
18) Do not output prose, comments, analysis, or markdown outside the YAML block.
19) The first characters must be exactly ```yaml and the final characters must be exactly ```.
\end{promptbox}

\begin{promptbox}{Orchestrator User Prompt Template}
Task type: [TASK_TYPE]
Difficulty: [DIFFICULTY]
Role pool variant: [ROLE_POOL_VARIANT]
Allowed base roles: [ALLOWED_BASE_ROLES]
Allowed final base roles: [FINAL_BASE_ROLES]

Problem:
[TASK]

Generate one complete YAML that decides custom agent names, base-role inheritance, topology wiring, and runtime model capacity_levels in one pass.
Return exactly one fenced YAML block: ```yaml ... ```.

Required output contract:
- First non-empty key must be defaults: or steps:
- Allowed keys only: defaults, steps, capacity_level, agents, type, base_role, duty, ref, active
- Do not include any extra fields such as name, description, model, prompt, output, or rationale
- Every active agent.type is a unique custom role id, not a fixed-pool base role
- Every active agent.base_role is one of the allowed base roles
- Every active agent.duty is non-empty and describes the node's local contribution
- The duty should be specific to the input problem and should state the expected local output
- Valid capacity_levels are exactly: small, medium, large
- Use small for cheaper/simple nodes, medium for normal reasoning, and large only when stronger reasoning is justified
- step[0] refs must be empty
- refs must point to active agent.type values from earlier steps only
- refs must not point to undefined, inactive, duplicate, current-step, or future agents
- final active step must include a custom final-answer agent using one of the allowed final base roles
- Keep the graph executable and concise, but do not force a fixed step count
- Avoid redundant agents that perform the same local duty
- Avoid unnecessary all-to-all references when a smaller dependency set is sufficient
- no prose, no analysis, no comments, no extra fields outside the YAML block

small/medium/large are runtime compute capacity_levels. 

Duty writing requirements:
- Use one concrete sentence per agent duty.
- Mention the local role intent and expected local output.
- Tie the duty to the given problem rather than writing a generic agent description.
- Avoid generic duties that do not identify what the node contributes.
- The final-answer duty should explicitly state the required benchmark answer format.

Example style:
Minimal topology example (schema only):
```yaml
defaults:
  capacity_level: medium
steps:
  - agents:
      - type: draft_agent
        base_role: [BASE_ROLE]
        duty: Extract the target information and produce a concise draft.
        ref: []
        capacity_level: medium
  - agents:
      - type: final_agent
        base_role: [FINAL_BASE_ROLE]
        duty: Check referenced work and return the benchmark answer format.
        ref: [draft_agent]
        capacity_level: medium
```

Now generate ONE valid full YAML for the target problem.
\end{promptbox}

\begin{promptbox}{YAML Repair Prompt}
You are a strict topology-and-budget YAML repairer.
Return exactly one YAML code block and nothing else.
Repair the provided full YAML so it satisfies inline custom-role, structure, ref, and capacity_level constraints.
Do not explain the fix.

Task type: [TASK_TYPE]
Difficulty: [DIFFICULTY]
Role pool variant: [ROLE_POOL_VARIANT]
Allowed base roles: [ALLOWED_BASE_ROLES]
Allowed final base roles: [FINAL_BASE_ROLES]

Problem:
[TASK]

Repair contract:
- Return exactly one fenced YAML block: ```yaml ... ```
- Repair inline custom-role fields, structure, refs, and capacity_level legality
- Preserve valid parts of the original YAML whenever possible
- Remove unsupported keys and illegal fields
- Every active agent must have a unique type, legal base_role, non-empty duty, ref, and capacity_level
- If an agent.type duplicates another active agent.type, rename it to a unique custom role id
- If an agent.base_role is not in the allowed base roles, replace it with the closest legal base role
- Rewrite short or generic duties so each one identifies the local output
- Ensure step[0] refs are empty
- Ensure all refs point only to active agent.type values introduced in earlier steps
- Remove refs to undefined, inactive, same-step, current, or future agents
- Ensure all capacity_level values are exactly one of: small, medium, large
- Add missing capacity_level fields using defaults, step-level, or agent-level capacity_level
- Use small for simple extraction or formatting nodes, medium for normal reasoning, and large only when clearly justified
- Ensure the final active step includes a custom final-answer agent using one of the allowed final base roles
- Keep the repaired graph executable and concise
- Do not add unnecessary steps, redundant agents, or avoidable dependency references

small/medium/large are runtime compute capacity_levels. 

Duty writing requirements:
- Use one concrete sentence per agent duty.
- Mention the local role intent and expected local output.
- Avoid generic duties that do not identify what the node contributes.
- The final-answer duty should explicitly state the required benchmark answer format.

Invalid full YAML to repair:
[INVALID_YAML]

Repair feedback:
[VALIDATION_FEEDBACK]
\end{promptbox}

\begin{promptbox}{Role-conditioned Worker Prompt Interface}
Role-conditioned worker instructions are dispatched by base_role.
The worker receives the instantiated custom role, inherited base role, task-specific duty, task input, referenced upstream outputs, and optional feedback.

Task type: [TASK_TYPE]
Derived role: [TYPE]
Inherited base role: [BASE_ROLE]
Extra duty: [DUTY]
Assigned capacity_level: [CAPACITY_LEVEL]

Problem:
[TASK]

Referenced upstream outputs:
[REFERENCED_OUTPUTS]

Feedback:
[FEEDBACK]

Worker instruction:
Follow the inherited role category and the task-specific duty for this problem.
Use referenced upstream outputs when they are relevant to the assigned duty.
Do not repeat upstream reasoning unless verification or correction is required by the duty.
Do not solve unrelated parts of the task outside the assigned responsibility.
Produce a concise output that can be consumed by downstream agents.
If the duty requires verification, explicitly state whether the referenced answer is accepted or corrected.
If the duty requires finalization, return the answer in the benchmark-required format.
\end{promptbox}

\section{Additional Experiments}
\label{app:additional_experiments}

\subsection{Training-Stage Ablations}
\label{app:training_stages}

Table~\ref{tab:training_stages_full} compares the complete training pipeline with supervised initialization alone and reinforcement learning without supervised initialization. The SFT-only, SFT + GRPO, and full variants start from the same supervised checkpoint, isolating the gains from orchestration-level GRPO and localized counterfactual credit assignment. SFT only reaches 72.89\% / 81.65\% / 79.53\% accuracy on MMLU / GSM8K / AQuA, below the strongest single-agent results of 74.46\% / 93.25\% / 82.68\% in Table~\ref{tab:main_results}. Subsequent reinforcement learning improves task accuracy, and adding counterfactual supervision further reduces worker-token usage. RL without SFT improves over the initial specification generator but remains below the full method, supporting supervised initialization as a useful starting point for execution-based learning. Full-evaluation-set specification validity is reported in Appendix~\ref{app:serialization_validity}.

\begin{table}[htbp]
\centering\small
\caption{Complete training-stage comparison. Acc. denotes task accuracy (\%); Tok. denotes average worker-token usage per task.}
\label{tab:training_stages_full}
\setlength{\tabcolsep}{4pt}
\begin{tabular}{lrrrrrr}
\toprule
& \multicolumn{2}{c}{MMLU} & \multicolumn{2}{c}{GSM8K} & \multicolumn{2}{c}{AQuA} \\
Setting & Acc. & Tok. & Acc. & Tok. & Acc. & Tok. \\
\midrule
Before SFT, no RL & 16.39 & 3443.61 & 46.24 & 3741.83 & 4.33 & 2902.22 \\
RL without SFT & 48.53 & 2777.60 & 75.82 & 3964.21 & 64.96 & 3729.20 \\
SFT only & 72.89 & 1950.51 & 81.65 & 1564.01 & 79.53 & 1421.90 \\
SFT + GRPO & 83.08 & 2543.21 & 93.40 & 3742.59 & 85.43 & 3012.54 \\
SFT + GRPO + CF & 84.32 & 2148.00 & 94.32 & 3362.00 & 86.22 & 2056.10 \\
\bottomrule
\end{tabular}
\end{table}

\noindent\textbf{Orchestration design ablations.}
Table~\ref{tab:design_ablation_settings} defines the component ablations reported in Table~\ref{tab:ablation}. Each variant removes one type of task-adaptive orchestration decision. The counterfactual mutation ablations separately disable dependency deletion, role rollback, or capacity downgrade to examine the contribution of each localized credit signal.

\begin{table}[htbp]
\centering\small
\caption{Definitions of the orchestration design ablations.}
\label{tab:design_ablation_settings}
\begin{tabular}{p{0.28\linewidth}p{0.64\linewidth}}
\toprule
Variant & Modification \\
\midrule
w/o Role Design & Replace each customized duty with the default duty associated with its inherited base role. \\
w/o Capacity Assignment & Assign the 14B worker to all LLM worker agents. \\
w/o Dependency Construction & Remove all inter-agent dependency references (\texttt{ref: []}), so workers receive no upstream agent outputs. \\
\bottomrule
\end{tabular}
\end{table}

\subsection{Sensitivity to Reward and Counterfactual Weights}
\label{app:weight_sensitivity}

The reward preference configuration was developed on GSM8K and then fixed for the other five main benchmarks, without benchmark-specific tuning. Table~\ref{tab:weight_sensitivity} varies one coefficient at a time while holding the others at their defaults: $\lambda_{\mathrm{tok}}=1.5$, $\lambda_{\mathrm{str}}=0.1$, and $\lambda_{\mathrm{cf}}=0.05$.

\begin{table}[htbp]
\centering\small
\caption{One-factor-at-a-time sensitivity. Accuracy and average worker tokens are reported in the order MMLU / GSM8K / AQuA.}
\label{tab:weight_sensitivity}
\setlength{\tabcolsep}{4pt}
\begin{tabular}{lcll}
\toprule
Coefficient & Value & Accuracy (\%) & Worker tokens \\
\midrule
$\lambda_{\mathrm{tok}}$ & 0.75 & 84.98 / 94.69 / 86.22 & 3212.19 / 4801.30 / 2964.02 \\
& 1.5 & 84.32 / 94.32 / 86.22 & 2148.00 / 3362.00 / 2056.10 \\
& 3.0 & 72.24 / 86.20 / 77.56 & 1645.08 / 2454.70 / 1443.78 \\
\midrule
$\lambda_{\mathrm{str}}$ & 0.05 & 84.25 / 93.93 / 85.82 & 2256.43 / 3585.36 / 2210.45 \\
& 0.1 & 84.32 / 94.32 / 86.22 & 2148.00 / 3362.00 / 2056.10 \\
& 0.2 & 80.27 / 92.64 / 83.46 & 1894.00 / 3204.30 / 1903.20 \\
\midrule
$\lambda_{\mathrm{cf}}$ & 0 & 83.08 / 93.40 / 85.43 & 2543.21 / 3742.59 / 3012.54 \\
& 0.05 & 84.32 / 94.32 / 86.22 & 2148.00 / 3362.00 / 2056.10 \\
& 0.10 & 78.38 / 87.33 / 77.56 & 2643.23 / 3421.03 / 2490.54 \\
\bottomrule
\end{tabular}
\end{table}

Increasing $\lambda_{\mathrm{tok}}$ from 0.75 to 1.5 reduces tokens by approximately 30--33\% with accuracy changes of at most 0.66 points; increasing it to 3.0 trades further token savings for larger accuracy losses. Changing $\lambda_{\mathrm{str}}$ from 0.05 to 0.1 changes accuracy by at most 0.40 points while reducing token usage. The default counterfactual coefficient improves both accuracy and token efficiency over $\lambda_{\mathrm{cf}}=0$, whereas 0.10 degrades accuracy. These results illustrate how the reward weights control the accuracy--token trade-off and how the counterfactual weight balances localized supervision with orchestration-level GRPO.

\noindent\textbf{Reward interpretation.}
Task correctness is evaluated externally, and cost is measured from worker execution and the compiled graph. The bounded efficiency bonus and the reduced bonus for unsolved specifications discourage cost reduction at the expense of correctness. We report task accuracy, worker-token usage, and specification validity separately to evaluate the learned orchestration beyond its shaped reward.

\subsection{Serialization and Specification Validity}
\label{app:serialization_validity}

YAML and JSON can encode the same orchestration fields and dependencies. Table~\ref{tab:serialization} compares the two representations on GSM8K. We train separate orchestrators for YAML and JSON. The accuracy difference is 0.24 percentage points, with similar worker-token usage and parse success, indicating that the effectiveness of the learned orchestration is preserved across these two serialization formats.

\begin{table}[htbp]
\centering\small
\caption{Serialization comparison on GSM8K. Parse success is reported separately from task accuracy.}
\label{tab:serialization}
\begin{tabular}{lrrr}
\toprule
Representation & Accuracy (\%) & Worker tokens & Parse success (\%) \\
\midrule
YAML & 94.32 & 3362.00 & 99.70 \\
JSON & 94.08 & 3305.42 & 99.31 \\
\bottomrule
\end{tabular}
\end{table}

Specification validity requires a parseable, field-complete object with legal base roles and capacities, backward-only references, a valid terminal role, and successful DAG compilation. It does not imply task success: valid but ineffective orchestrations are evaluated using the same execution reward. Table~\ref{tab:full_validity} reports specification validity across the full evaluation sets, with rates of 99.40\% or higher on all six benchmarks. Parse success in Table~\ref{tab:serialization} checks whether an output can be parsed, whereas specification validity additionally requires the execution constraints above.

\begin{table}[htbp]
\centering\small
\caption{Specification validity across the full evaluation sets.}
\label{tab:full_validity}
\begin{tabular}{lrrr}
\toprule
Dataset & Evaluated & Valid & Rate (\%) \\
\midrule
MMLU (val) & 1531 & 1527 & 99.74 \\
GSM8K (test) & 1319 & 1315 & 99.70 \\
AQuA (test) & 254 & 254 & 100.00 \\
MultiArith & 600 & 598 & 99.67 \\
SVAMP & 1000 & 994 & 99.40 \\
HumanEval (held-out) & 124 & 124 & 100.00 \\
\bottomrule
\end{tabular}
\end{table}

\section{Local First-Order Analysis of Counterfactual Learning}
\label{app:cf_local_analysis}

We analyze one fixed retained pair $(y,\widetilde y)$ with observed reward contrast $|\Delta R|\geq\delta_{\min}>0$. The sampled specifications, token spans, observed rewards, and contrast weight are held fixed during the update. Define the preferred score margin
\begin{equation}
 m_\theta=\operatorname{sign}(\Delta R)
 \bigl(s_\theta^{\mathrm{fact}}-s_\theta^{\mathrm{cf}}\bigr),
 \qquad
 \ell_{\mathrm{pair}}(\theta)=-w_{\mathrm{cf}}(\Delta R)
 \log\operatorname{sigmoid}(\beta_{\mathrm{cf}}m_\theta).
\end{equation}

\noindent\textbf{Local margin improvement.}
Assume $m_\theta$ is twice continuously differentiable in a neighborhood of the current parameters. For an isolated gradient-descent step $\theta^+=\theta-\eta\nabla_\theta\ell_{\mathrm{pair}}(\theta)$, with sufficiently small $\eta>0$,
\begin{equation}
 m_{\theta^+}-m_\theta
 =\eta a_\theta\|\nabla_\theta m_\theta\|_2^2+O(\eta^2),
 \qquad
 a_\theta=w_{\mathrm{cf}}(\Delta R)\beta_{\mathrm{cf}}
 \operatorname{sigmoid}(-\beta_{\mathrm{cf}}m_\theta)>0.
 \label{eq:cf_margin_improvement}
\end{equation}
Thus the first-order change is nonnegative and is strictly positive when the margin gradient is nonzero.

\noindent\textbf{Derivation.}
Differentiating the pairwise loss gives $\nabla_\theta\ell_{\mathrm{pair}}=-a_\theta\nabla_\theta m_\theta$. Substituting $\theta^+-\theta=\eta a_\theta\nabla_\theta m_\theta$ into the first-order Taylor expansion of $m_{\theta^+}$ gives Eq.~\ref{eq:cf_margin_improvement}.

This result explains how the counterfactual loss favors the field value with the higher observed reward through a local update on the selected pair.

\section{Role Interfaces and Worker Execution}
\label{app:role_interfaces}
\label{app:implementation_notes}

The fixed schema and runtime interfaces constrain executability, while the policy generates the YAML instance and its orchestration graph for each task. A generated \texttt{type} and \texttt{duty} specify a task-specific identity and responsibility; \texttt{base\_role} selects an execution interface. Capacity chooses the worker backbone, and dependency references specify upstream context. Table~\ref{tab:role_interfaces} lists the domain-specific interface pools.

\begin{table}[htbp]
\centering\small
\caption{Domain-specific base-role interfaces. Task-specific duties are generated rather than selected from this list.}
\label{tab:role_interfaces}
\begin{tabular}{p{0.24\linewidth}p{0.69\linewidth}}
\toprule
Domain & Interfaces \\
\midrule
GSM8K / MultiArith / SVAMP & \texttt{math\_solver}, \texttt{math\_analyst}, \texttt{python\_solver}, \texttt{solution\_inspector}, \texttt{final\_decider}, \texttt{majority\_voter} \\
AQuA & \texttt{choice\_math\_solver}, \texttt{choice\_analyst}, \texttt{choice\_python\_solver}, \texttt{choice\_inspector}, \texttt{choice\_final\_decider}, \texttt{majority\_voter} \\
MMLU & \texttt{knowledge\_expert}, \texttt{domain\_specialist}, \texttt{option\_analyst}, \texttt{knowledge\_critic}, \texttt{choice\_final\_decider}, \texttt{majority\_voter} \\
HumanEval & \texttt{code\_planner}, \texttt{algorithm\_designer}, \texttt{code\_writer}, \texttt{test\_designer}, \texttt{bug\_fixer}, \texttt{final\_code\_writer} \\
\bottomrule
\end{tabular}
\end{table}

Most interfaces instantiate an LLM worker; \texttt{majority\_voter} performs deterministic aggregation. Tool availability is bound to compatible interfaces rather than freely selected through generated duties: \texttt{python\_exec} serves \texttt{python\_solver} and \texttt{choice\_python\_solver}; \texttt{unit\_test\_runner} serves \texttt{final\_code\_writer}; and \texttt{answer\_parser} supports final-answer interfaces and voting. The \texttt{math\_calc} interfaces are \texttt{math\_solver}, \texttt{math\_analyst}, \texttt{python\_solver}, \texttt{choice\_math\_solver}, and \texttt{choice\_python\_solver}.

The runtime constructs each worker prompt from its inherited base role, task-specific duty, assigned capacity, task input, and referenced upstream outputs. The validated specification determines the resulting execution structure.

\section{Limitations}
\label{app:limitations}
\methodname{} has two main limitations. First, training requires repeated worker execution to evaluate sampled specifications and their counterfactual variants. Node-level caching reduces redundant computation, but execution-based reinforcement learning still incurs substantial training cost. Second, our evaluation focuses on text-based reasoning and coding benchmarks. Extending compositional orchestration to long-horizon tool use, multimodal inputs, and open-ended interactive environments remains an important direction for future work.

\section{Counterfactual Overhead and Execution Caching}
\label{app:cf_overhead}

\begin{figure}[htbp]
    \centering
    \includegraphics[width=0.60\linewidth]{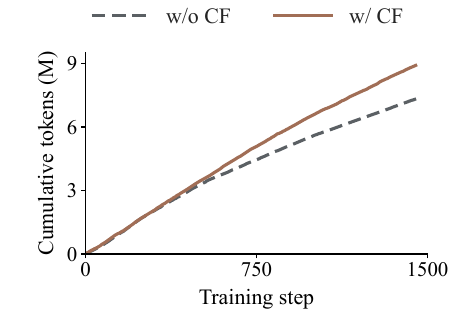}
    \caption{Recorded cumulative training-token usage with and without localized counterfactual credit assignment, in millions of tokens.}
    \label{fig:cf_training_cost}
\end{figure}

Figure~\ref{fig:cf_training_cost} compares recorded cumulative training-token usage with and without localized counterfactual credit assignment. Training-token usage is accumulated using the same capacity weights as the worker-token metric in Appendix~\ref{eval_metric}. The full method uses approximately 8.90M recorded tokens, compared with 7.33M for orchestration-level GRPO alone, an increase of 21.6\%. Together with Figure~\ref{fig:cf_analyse}, these results illustrate a training--execution trade-off: the additional training computation accompanies a higher final orchestration reward and an 11.1\% reduction in worker-token usage per execution. Counterfactual comparisons therefore increase the training investment while helping the orchestrator learn more effective and token-efficient specifications.

For each valid rollout, we sample at most one feasible mutation family and one edit location to construct a counterfactual specification. Evaluating this specification requires additional worker executions. To reduce redundant computation, we use node-level caching keyed by the input task, base role, customized duty, capacity level, and referenced upstream outputs. Worker outputs are reused whenever these inputs remain unchanged.

A local edit requires recomputation only for worker calls whose prompts or upstream contexts change. Dependency deletion affects the receiving agent and its downstream dependents, while role rollback and capacity downgrade affect the edited agent and its downstream dependents. Unrelated branches remain reusable, concentrating the additional execution on the affected portion of the orchestration graph.

Counterfactual evaluation is used only during reinforcement learning. At inference time, \methodname{} generates and executes one orchestration specification per task, so the learned improvements require no counterfactual comparisons during deployment.

\section{Reproducibility Details}
\label{app:reproducibility}

\subsection{Model Deployment}
\label{app:model_deployment}

Table~\ref{tab:model_deployment} reports the model backbones used by the orchestration policy and worker agents. The orchestration policy is initialized from a 7B instruction model and is trained to generate executable specifications. During execution, the generated capacity level selects one of three worker backbones. The small, medium, and large capacity levels correspond to increasingly stronger Qwen2.5-Instruct models.

\begin{table}[h]
\centering
\caption{Model deployment used in our experiments.}
\label{tab:model_deployment}
\small
\begin{tabular}{ll}
\toprule
\textbf{Role} & \textbf{Model} \\
\midrule
Orchestration policy backbone & \texttt{Qwen/Qwen2.5-7B-Instruct} \\
Small-capacity worker & \texttt{Qwen/Qwen2.5-7B-Instruct} \\
Medium-capacity worker & \texttt{Qwen/Qwen2.5-14B-Instruct} \\
Large-capacity worker & \texttt{Qwen/Qwen2.5-32B-Instruct} \\
\bottomrule
\end{tabular}
\end{table}

This deployment allows the orchestrator to allocate different worker capacities within the same generated specification, rather than using a homogeneous worker model for all agents.

\subsection{Main Training Hyperparameters}
\label{app:main_hyperparams}

Table~\ref{tab:main_hyperparams} summarizes the main optimization and generation hyperparameters used for reinforcement learning. We train for 1500 steps with at most 480 RL training samples per dataset. For each task, the policy samples a group of candidate orchestration specifications, executes valid candidates, and uses group-relative rewards for policy optimization. The prompt and generation length limits are set to 2048 tokens to allow the policy to output complete YAML specifications while keeping rollout cost bounded.

\begin{table}[h]
\centering
\caption{Main training hyperparameters.}
\label{tab:main_hyperparams}
\small
\begin{tabular}{ll}
\toprule
\textbf{Hyperparameter} & \textbf{Value} \\
\midrule
Training-dynamics seeds & 42, 1701, 67, 0 \\
Training steps & 1500 \\
Max train samples & 480 \\
Split preset & \texttt{cross\_dataset\_v1} \\
Group size / generations & 4 \\
Prompts per reward call & 4 \\
Candidate rows per reward call & 16 \\
Learning rate & $1\times 10^{-6}$ \\
LoRA & enabled \\
LoRA rank / alpha & 8 / 32 \\
Gradient checkpointing & enabled \\
Train batch size & 4 \\
Gradient accumulation & 1 \\
Max prompt tokens & 2048 \\
Max new tokens & 2048 \\
Temperature / top-$p$ & 0.6 / 0.9 \\
Role pool variant & \texttt{baseline} \\
Node cache & enabled \\
\bottomrule
\end{tabular}
\end{table}

Each reward call evaluates four tasks with four sampled specifications per task, yielding sixteen candidates. LoRA and gradient checkpointing reduce memory usage during policy optimization; execution caching is described in Appendix~\ref{app:cf_overhead}.

\noindent\textbf{Training-dynamics statistics.}
The training-dynamics comparison uses four random seeds: 42, 1701, 67, and 0. Shaded regions in Figure~\ref{fig:cf_analyse}(a,b) indicate $\pm 1$ standard deviation across these runs.

\subsection{GRPO Objective}
\label{app:grpo_objective}

Let $\mathcal{D}_{\mathrm{train}}$ denote the training dataset, from which tasks are sampled. For each task $x$, the rollout policy samples $K$ specifications $y_i\sim\pi_{\theta_{\mathrm{old}}}(\cdot\mid x)$, for $i=1,\ldots,K$, and assigns each specification the reward $R_i=R_{\mathrm{orch}}(x,y_i)$. Define the group mean, standard deviation, and normalized advantage as
\begin{equation}
\begin{aligned}
    \overline{R}_x&=\frac{1}{K}\sum_{i=1}^{K}R_i,
    &\sigma_x&=\sqrt{\frac{1}{K}\sum_{i=1}^{K}(R_i-\overline{R}_x)^2},\\
    \widehat{A}_i&=\frac{R_i-\overline{R}_x}{\sigma_x+\epsilon_{\mathrm{adv}}},
\end{aligned}
\label{eq:group_advantage}
\end{equation}
where $\epsilon_{\mathrm{adv}}>0$ ensures numerical stability. Following standard GRPO~\cite{guo2025deepseek}, the token-wise importance ratio is
\begin{equation}
    r_{i,t}(\theta)=
    \frac{\pi_{\theta}(y_{i,t}\mid x,y_{i,<t})}
    {\pi_{\theta_{\mathrm{old}}}(y_{i,t}\mid x,y_{i,<t})},
    \label{eq:token_ratio}
\end{equation}
where $y_{i,t}$ is token $t$ of the serialized specification $y_i$ and $y_{i,<t}$ is its preceding token sequence. The clipped surrogate in Eq.~\ref{eq:grpo_loss} is
\begin{equation}
\ell_{\mathrm{clip}}(r,A)
=\min\!\left(rA,\operatorname{clip}(r,1-\epsilon_{\mathrm{clip}},1+\epsilon_{\mathrm{clip}})A\right),
\end{equation}
where $\epsilon_{\mathrm{clip}}$ is the clipping coefficient.

\subsection{Reward Hyperparameters}
\label{app:reward_hyperparams}

Table~\ref{tab:reward_hyperparams} reports the reward weights used throughout the main experiments. Appendix~\ref{app:weight_sensitivity} examines their effect on accuracy and worker-token usage.

\begin{table}[h]
\centering
\caption{Reward hyperparameters.}
\label{tab:reward_hyperparams}
\small
\begin{tabular}{lr}
\toprule
\textbf{Hyperparameter} & \textbf{Value} \\
\midrule
\texttt{efficiency\_weight} ($\lambda_{\mathrm{tok}}$) & 1.5 \\
\texttt{structure\_weight} ($\lambda_{\mathrm{str}}$) & 0.1 \\
\texttt{execution\_weight} & 1.0 \\
\bottomrule
\end{tabular}
\end{table}

\subsection{Localized Counterfactual Training}
\label{app:cf_hyperparams}

Table~\ref{tab:cf_hyperparams} lists the settings for counterfactual sampling and optimization. Each pair changes one dependency, role, or capacity field, as defined in Section~\ref{sec:rl}.

\begin{table}[h]
\centering
\caption{Localized counterfactual training hyperparameters.}
\label{tab:cf_hyperparams}
\small
\begin{tabular}{ll}
\toprule
\textbf{Hyperparameter} & \textbf{Value} \\
\midrule
Localized counterfactual objective & enabled \\
Mutation families & dependency deletion, role rollback, capacity downgrade \\
Dependency deletion count & 1 \\
Role rollback count & 1 \\
Capacity downgrade count & 1 \\
Sampling mode & \texttt{adaptive\_rotation} \\
Reward contrast in the CF loss & Raw $\Delta R$ (Eq.~\ref{eq:cf_delta}) \\
EMA alpha & 0.1 \\
Rotation temperature & 1.0 \\
Minimum type probability & 0.05 \\
Counterfactual objective weight ($\lambda_{\mathrm{cf}}$) & 0.05 \\
Preference beta ($\beta_{\mathrm{cf}}$) & 0.1 \\
Delta cap ($\delta_{\max}$) & 0.5 \\
Minimum absolute delta ($\delta_{\min}$) & 0.01 \\
\bottomrule
\end{tabular}
\end{table}

The mutation counts are set to one so that each counterfactual pair changes only one local orchestration field. This keeps the reward contrast attributable to a specific dependency, role, or capacity decision. The adaptive rotation sampler adjusts the probability of choosing each mutation family according to its recent reward contrasts, while the minimum type probability preserves exploration over all mutation types. The delta cap and minimum absolute delta stabilize the counterfactual signal by clipping very large contrasts and filtering negligible reward differences. The counterfactual objective weight controls the contribution of localized counterfactual learning relative to the orchestration-level GRPO objective.

\noindent\textbf{Field-span construction.}
The factual and counterfactual scoring spans correspond to the complete serialized values of the selected field. For dependency deletion, both spans include the brackets of the \texttt{ref} list: deleting its only reference compares \texttt{[A]} with \texttt{[]}. The empty list therefore retains a nonempty token span, so the average log-probability remains well-defined. Tokens outside the selected field values provide conditioning context but are excluded from the span score.

\noindent\textbf{Reward contrast.}
The localized loss uses the raw orchestration-reward difference $\Delta R$ in Eq.~\ref{eq:cf_delta}. Its sign determines the preference direction, while $w_{\mathrm{cf}}(\Delta R)$ and $\delta_{\min}$ control the contrast weight and filtering condition, respectively.

\subsection{Supervised Warm Start}
\label{app:sft_warm_start}

We construct 500 supervised examples from questions randomly selected from the training set, using a 32B teacher to generate executable YAML specifications. These examples initialize the orchestrator's ability to generate role, capacity, and dependency fields before execution-based reinforcement learning. We fine-tune the orchestration policy with LoRA using LLaMA-Factory for 32 global steps and use the resulting checkpoint to initialize GRPO training. The training-stage comparison is reported in Appendix~\ref{app:training_stages}.

\subsection{Evaluation Protocol}
\label{app:evaluation_protocol}

For RL training, we sample at most 480 examples from the training split of each dataset when an official training split is available. For HumanEval, which does not provide a separate training split, we use 40 examples for RL training or task-level adaptation by methods that require it; all methods are evaluated on the remaining 124. For MMLU, we evaluate on the validation split because the official test split is substantially larger and more costly.

Evaluation is conducted on MMLU, GSM8K, AQuA, MultiArith, SVAMP, and HumanEval. For each evaluation instance, the orchestrator generates one executable orchestration specification. The validator checks schema validity and dependency validity, and the compiled orchestration graph is executed once to produce the final answer. Table~\ref{tab:evaluation_datasets} summarizes the evaluation split and task type for each benchmark.

\begin{table}[h]
\centering
\caption{Evaluation datasets.}
\label{tab:evaluation_datasets}
\small
\begin{tabular}{lrl}
\toprule
\textbf{Dataset} & \textbf{Evaluation size} & \textbf{Task type} \\
\midrule
MMLU (val) & 1531 & multiple-choice reasoning \\
GSM8K (test) & 1319 & mathematical reasoning \\
AQuA (test) & 254 & mathematical reasoning \\
MultiArith & 600 & mathematical reasoning \\
SVAMP & 1000 & mathematical reasoning \\
HumanEval (held-out) & 124 & code generation \\
\bottomrule
\end{tabular}
\end{table}

Evaluation metrics are defined in Appendix~\ref{eval_metric}. The 124 held-out HumanEval examples do not overlap with the 40 examples used for RL training or adaptation.

\section{Compute Resources}
\label{app:compute}

Table~\ref{tab:compute} reports the compute resources used for training and evaluation. The main RL run takes 24 h 50 min; the counterfactual execution and caching procedure is described in Appendix~\ref{app:cf_overhead}.

\begin{table}[h]
\centering
\caption{Compute resources used in our experiments.}
\label{tab:compute}
\small
\begin{tabular}{lc}
\toprule
\textbf{Item} & \textbf{Value} \\
\midrule
Compute platform & Amazon EC2 \\
Instance type & \texttt{g6e.12xlarge} \\
GPU type & NVIDIA L40S \\
Number of GPUs & 4 \\
GPU memory & 4 $\times$ 48 GB \\
CPU type & AMD EPYC 7R13 \\
vCPUs & 48 \\
System memory & 384 GiB \\
Storage used & 1000 GB \\
Main RL training time & 24 h 50 min \\
\bottomrule
\end{tabular}
\end{table}

\section{Broader Impacts}
\label{app:broader_impacts}
\methodname{} aims to improve the effectiveness and efficiency of LLM-based multi-agent orchestration. Its potential positive impacts include reducing redundant worker-model calls, improving the efficiency of multi-agent reasoning systems, and enabling more systematic construction of task-adaptive orchestration specifications. These properties may make multi-agent systems easier to study, compare, and deploy under controlled resource budgets. As with other LLM-based systems, practical deployment should still include task-specific validation and appropriate human oversight.

\section{Existing Assets and Licenses}
\label{app:licenses}

We use existing datasets, baseline implementations, and LLM backbones for research evaluation. We cite the original papers or repositories for these assets and follow their licenses or terms of use. For assets whose license is not explicitly specified in the local snapshot, we report the license status as not found and use the asset only for research comparison without redistributing it. Table~\ref{tab:existing_assets} summarizes the main existing assets used in this work.

\begin{table}[h]
\centering
\caption{Existing assets used in this work.}
\label{tab:existing_assets}
\small
\resizebox{\textwidth}{!}{
\begin{tabular}{llll}
\toprule
\textbf{Asset} & \textbf{Type} & \textbf{Source / Version} & \textbf{License / Terms} \\
\midrule
MMLU & Dataset & Hendrycks et al.; local snapshot & MIT License \\
GSM8K & Dataset & Cobbe et al.; local snapshot & MIT License \\
AQuA & Dataset & Ling et al.; local snapshot & Apache License 2.0 \\
MultiArith & Dataset & Roy and Roth; local snapshot & Not explicitly specified in local snapshot; research evaluation only \\
SVAMP & Dataset & Patel et al.; local snapshot & MIT License \\
HumanEval & Dataset & Chen et al.; local snapshot & MIT License \\
\midrule
AFlow & Baseline / Code & Original repository; local snapshot & MIT License \\
G-Designer & Baseline / Code & Original repository; local snapshot & License not found in local snapshot; research evaluation only \\
OFA-MAS & Baseline / Code & Original repository; local snapshot & License not found in local snapshot; research evaluation only \\
ARG-DESIGNER & Baseline / Code & Original repository; local snapshot & License not found in local snapshot; research evaluation only \\
\midrule
Qwen2.5-7B-Instruct & Model & \texttt{Qwen/Qwen2.5-7B-Instruct} & Apache License 2.0 \\
Qwen2.5-14B-Instruct & Model & \texttt{Qwen/Qwen2.5-14B-Instruct} & Apache License 2.0 \\
Qwen2.5-32B-Instruct & Model & \texttt{Qwen/Qwen2.5-32B-Instruct} & Apache License 2.0 \\
\bottomrule
\end{tabular}
}
\end{table}

\end{document}